\documentclass{article}
\usepackage{arxiv}
\usepackage[utf8]{inputenc} 
\usepackage[T1]{fontenc}    
\usepackage{hyperref}       
\usepackage{url}            
\usepackage{booktabs}       
\usepackage{amsfonts}       
\usepackage{nicefrac}       
\usepackage{microtype}      
\usepackage{lipsum}		
\usepackage{graphicx}
\usepackage{natbib}
\usepackage{doi}

\usepackage{lineno}
\usepackage{physics}
\usepackage{multirow}
\usepackage{caption,subcaption}

\usepackage{algorithm}
\usepackage{algorithmic}

\usepackage{hyperref}

\usepackage[flushleft]{threeparttable}

\usepackage[acronym,nomain]{glossaries}
\usepackage{amsmath}
\usepackage{amssymb}

\hypersetup{
    colorlinks=true,
    linkcolor=blue,
    filecolor=magenta,      
    urlcolor=cyan,
}

\makeglossaries

\newacronym{aa}{AA}{AutoAugment}
\newacronym{cnn}{CNN}{Convolutional Neural Network}
\newacronym{nas}{NAS}{Network Architecture Search}
\newacronym{co}{CO}{Cutout}
\newacronym{ea}{EA}{Evolutionary Algorithm}
\newacronym{eanet}{EA-Net}{Evolutionary Algorithm Network}
\newacronym{eeea}{EEEA}{Early Exit Evolutionary Algorithm}

\newacronym{eeeanet}{EEEA-Net}{Early Exit Evolutionary Algorithm Network}

\newacronym{eepi}{EE-PI}{Early Exit Population Initialisation}
\newacronym{flops}{FLOPS}{Floating point Operations Per Second}
\newacronym{ga}{GA}{Genetic Algorithm}
\newacronym{imagenet}{ImageNet}{ImageNet Large Scale Visual Recognition Challenge}
\newacronym{irn}{IRN}{Inverted Residuals Network}
\newacronym{rl}{RL}{Reinforcement Learning}
\newacronym{rn}{RN}{Residuals Network}
\newacronym{yolov4}{YOLOv4}{You Only Look Once version four}
\newacronym{miou}{mIoU}{Mean Intersection over Union}
\newacronym{ap}{AP}{Average Precision}
\newacronym{ssd}{SSD}{Single-Shot Detectors}

\title{EEEA-Net: An Early Exit Evolutionary Neural Architecture Search}

\date{} 					

\author{Chakkrit~Termritthikun\thanks{This work is done when Chakkrit Termritthikun works as a visiting research student in University of South Australia} \\
	STEM, University of South Australia\\
	Adelaide, SA, 5095, Australia\\
	\texttt{chakkritt60@nu.ac.th} \\
	\And
	Yeshi~Jamtsho \\
	College of Science and Technology\\
	Royal University of Bhutan\\
	Phuentsholing, 21101, Bhutan \\
	\texttt{yjamtsho.cst@rub.edu.bt} \\
	\And
	Jirarat~Ieamsaard \\
	Department of Electrical and Computer Engineering\\
	Faculty of Engineering, Naresuan University\\
	Phitsanulok, 65000, Thailand \\
	\texttt{jirarati@nu.ac.th} \\
	\And
	Paisarn~Muneesawang \\
	Department of Electrical and Computer Engineering\\
	Faculty of Engineering, Naresuan University\\
	Phitsanulok, 65000, Thailand \\
	\texttt{paisarnmu@nu.ac.th} \\
	\And
	Ivan~Lee \\
	STEM, University of South Australia\\
	Adelaide, SA, 5095, Australia\\
	\texttt{ivan.lee@unisa.edu.au} \\
	
}

\renewcommand{\headeright}{}
\renewcommand{\undertitle}{Published at Engineering Applications of Artificial Intelligence. DOI: \url{https://doi.org/10.1016/j.engappai.2021.104397}}
\renewcommand{\shorttitle}{EEEA: Early Exit Evolutionary Algorithm Network by Chakkrit Termritthikun, et al.}

\begin{document}

\maketitle

\begin{abstract}

The goals of this research were to search for \gls*{cnn} architectures, suitable for an on-device processor with limited computing resources, performing at substantially lower \gls*{nas} costs. A new algorithm entitled an \gls*{eepi} for \gls*{ea} was developed to achieve both goals. The \gls*{eepi} reduces the total number of parameters in the search process by filtering the models with fewer parameters than the maximum threshold. It will look for a new model to replace those models with parameters more than the threshold. Thereby, reducing the number of parameters, memory usage for model storage and processing time while maintaining the same performance or accuracy. The search time was reduced to 0.52 GPU day. This is a huge and significant achievement compared to the \gls*{nas} of 4 GPU days achieved using NSGA-Net, 3,150 GPU days by the AmoebaNet model, and the 2,000 GPU days by the NASNet model. As well, Early Exit Evolutionary Algorithm networks (EEEA-Nets) yield network architectures with minimal error and computational cost suitable for a given dataset as a class of network algorithms. Using EEEA-Net on CIFAR-10, CIFAR-100, and ImageNet datasets, our experiments showed that EEEA-Net achieved the lowest error rate among state-of-the-art \gls*{nas} models, with 2.46\% for CIFAR-10, 15.02\% for CIFAR-100, and 23.8\% for ImageNet dataset. Further, we implemented this image recognition architecture for other tasks, such as object detection, semantic segmentation, and keypoint detection tasks, and, in our experiments, EEEA-Net-C2 outperformed MobileNet-V3 on all of these various tasks. (The algorithm code is available at \url{https://github.com/chakkritte/EEEA-Net}).

\end{abstract}


\section{Introduction}

Deep convolutional neural networks (CNNs) have been widely used in computer vision applications, including image recognition, image detection, and image segmentation. In the ImageNet Large Scale Visual Recognition Challenge (ILSVRC) \cite{russakovsky2015imagenet}, the AlexNet \cite{krizhevsky2017imagenet}, GoogLeNet \cite{szegedy2015going}, ResNet \cite{he2016deep}, and SENet \cite{hu2018squeeze} were represented models that had been widely used in various applications. The SqueezeNet \cite{iandola2016squeezenet}, MobileNets \cite{howard2017mobilenets}, NUF-Net \cite{termritthikun2019on,termritthikun2020an}, and ShuffleNet \cite{zhang2018shufflenet} models were simultaneously developed to be used on devices with limited resources. All of these architecture networks have been strengthened and advanced by developers for many years.  

However, a significant drawback of the useability of these models, and to the development of the efficient CNNs models, was the dependence on the designer's expertise and experience, including utilising resources such as high-performance computing (HPC) for the experimentation. The datasets used for analysis also affect the model efficiency, depending on the different features that different datasets manifest, and all image recognition datasets require specialised research knowledge for each dataset when modelling. One algorithm, the \gls*{nas} \cite{zoph2016neural} was designed to search the \gls*{cnn} network architecture for different datasets and thereby avoided the hitherto human intervention or design activity except during the definition of the initial hyper-parameters.

The \gls*{cnn} network architecture is flexible, allowing the model to be developed in different structures, with the module structure consisting of layers, linked in sequence with different parameters. Models obtained from \gls*{nas} methods differ in structure and parameters, making \gls*{nas} searches more efficient in finding dataset models, resulting in finding a model for each particular dataset which have a unique structure and parameter set.

\gls*{rl} and gradient descent algorithms, automate the search for models of deep learning. Searching a model with an \gls*{rl} algorithm takes 2,000 GPU days to uncover an effective model. However, when the gradient descent algorithm is focused on searching a model with the highest accuracy for only one objective; it takes 4 GPU days. Both algorithms have difficulty in dealing with a multi-objective problem. \gls*{ea}, however, can solve multi-objective optimisation problems.

EA apply optimisation methods that mimic the evolution of living organisms in nature, including reproduction, mutation, recombination, and selection. \gls*{ea} can find the most suitable candidate solution with quality function rates. EA-based \gls*{nas} approaches are very robust with shallow error values for experiments with CIFAR-10 and CIFAR-100 datasets. However, past search algorithms in models of the EA-based \gls*{nas} approaches have taken up to 3,150 GPU days in \cite{real2019regularized}, 300 GPU days in \cite{liu2018hierarchical}, and 4 GPU days in \cite{liu2018darts}. Many network architectures built from \gls*{nas} have a high number of parameters and high computing costs. It is obvious that extensive network architectures must be avoided when attempting to identify a network architecture that is suitable for other applications.

A network architecture, built from the DARTS \cite{liu2018darts} search space, is a multi-path \gls*{nas}. However, excessive path-level selection is a problem for multi-path NASs where every operation of the super network of a multi-path \gls*{nas} takes a separate path. This means that all connected weights across all paths require considerable memory. 

The single-path \gls*{nas} \cite{stamoulis2019single} was introduced to solve the \gls*{nas} problem by finding a subset of kernel weights in the single layer. This layer is called a super kernel. We called the model built from the super kernel, the Supernet. Based on the Supernet, which has many subnets, whole subnets are trained at the same time through weight sharing. The Supernet can be sampled and deployed without re-training. 

In our current work, we developed the \gls*{eepi} method for Evolutionary NASs. \gls*{eepi} was applied to the Multi-Objective Evolutionary Algorithm (MOEA) in the first generation to locate the newly created model with a certain number of parameters, discarding models with parameters more than the set criteria. This process iteratively continues until the model with fewer parameters than the initial set criteria is found. Thus, \gls*{eepi} complements MOEA. We created this add-on to prevent models with a high number of parameters and a high number of \gls*{flops}. Also, using a small number of parameters helps reduce model search costs. 

The key contributions of this paper are:
\begin{itemize}
\item We introduce Evolutionary Neural Architecture Search (EA-Net), which adopts a multi-objective evolutionary algorithm for neural architecture search with three objectives: minimisation of errors, minimal number of parameters, and lowest computing costs.
\item We proposed a simple method called Early Exit Population Initialisation (EE-PI) to avoid a model with a high number of parameters and high computational cost, by filtering the neural network architecture based on the number of parameters in the first-generation of the evolution. The architectures obtained by this method are called Early Exit Evolutionary Algorithm networks (EEEA-Net).
\item We conduct extensive experiments to evaluate the effectiveness of an EEEA-Net by outperforming MobileNet-V3 for all of the image recognition, object detection, semantic segmentation, and keypoint detection. Also, the EEEA-Net was widely tested on standard CIFAR-10 \cite{krizhevsky2009learning}, CIFAR-100 \cite{krizhevsky2009learning}, ImageNet \cite{russakovsky2015imagenet}, PASCAL VOC \cite{everingham2010the}, Cityscapes \cite{cordts2016the}, and MS COCO \cite{lin2014microsoft} datasets. 
\end{itemize}

\section{Related Work}
\gls*{nas} was designed to search and design the model structure that suits the best to the applied dataset. Thus, the model obtained by \gls*{nas} has a small number of parameters with high performance. \gls*{nas} can find models suitable for both small and large datasets. The \gls*{nas} can be of single-objective \gls*{nas} and multi-objective \gls*{nas}: a single-objective \gls*{nas} considers models from a single objective such as error rate, number of parameters, or \gls*{flops}. The multi-objective \gls*{nas} considers models considering more than one objective, which we have adopted in this paper to optimise the model performance. 

\subsection{Network Architecture Search}

Setting optimised parameters in each layer, such as kernel size, kernel scroll position (stride), zero paddings, as well as the output size, is the main challenge in creating \gls*{cnn} architectures efficiently for a given dataset. The total parameters are directly proportional to the number of layers. Manually designing a model takes too long and requires considerable experimentation to achieve optimal performance, which is why an automated model discovery is essential. 

\begin{figure}[t!]
\centering\includegraphics[width=0.7\linewidth]{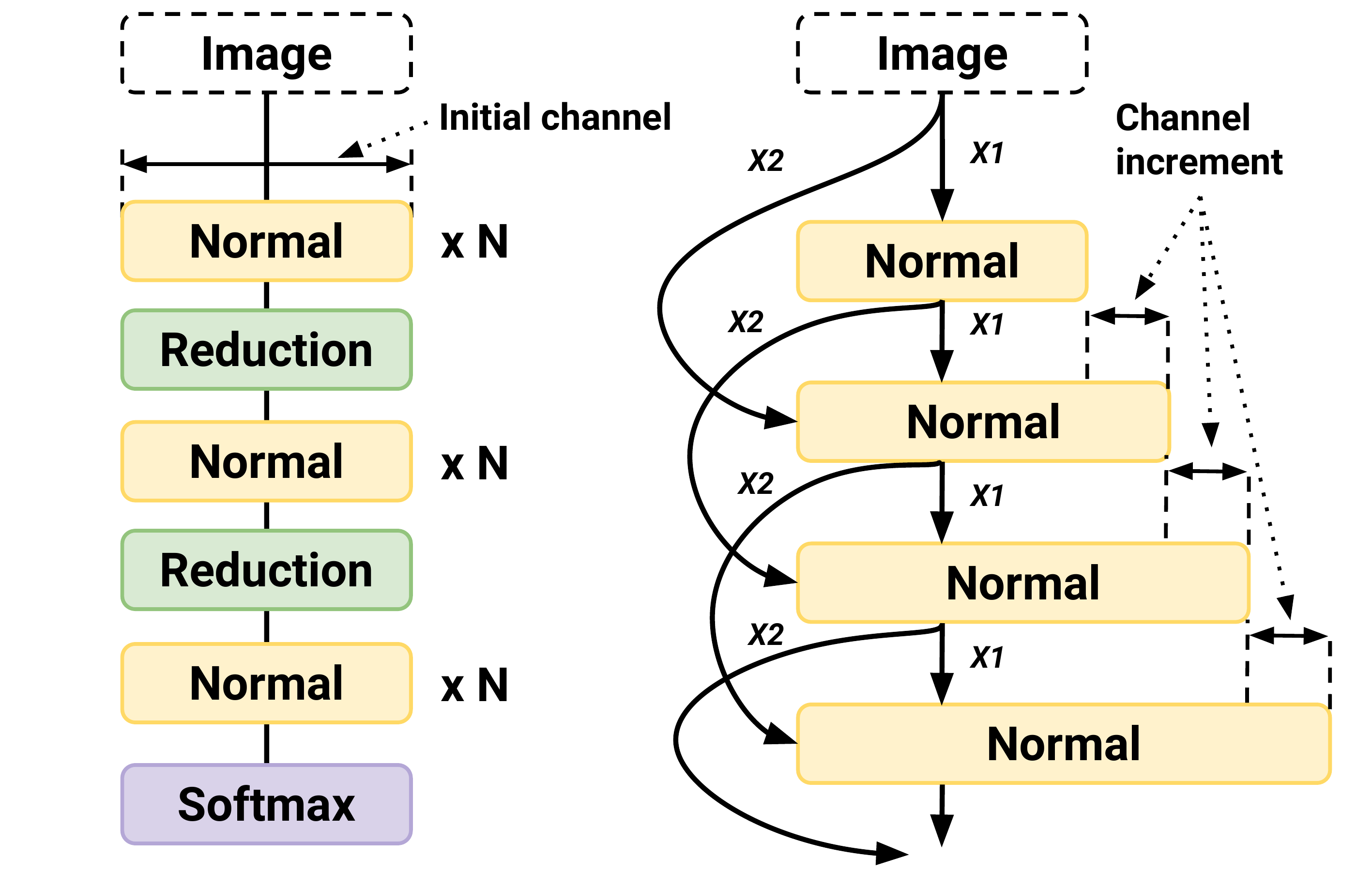}
\caption{NASNet \cite{zoph2016neural} network architecture (left) and NSGA-Net \cite{lu2019nsga} network architecture (right).}
\label{fig:1}
\end{figure}

The ability of NASs to find automated models suitable for datasets has proved a popular area of experimentation. Deep learning also is gaining popularity and is now being widely used. The \gls*{nas} model has also been designed and expanded to enable applications in new tasks such as a \gls*{nas} for semantic image segmentation, \gls*{nas} for object detection, and \gls*{nas} for skin lesion classification. The approaches used in \gls*{nas} are \gls*{rl} \cite{zoph2016neural}, \gls*{ea} \cite{real2019regularized,liu2018hierarchical,lu2019nsga}, and relaxation \cite{liu2018darts}. The three critical steps for \gls*{nas} are:

Step 1: Search the \gls*{cnn} model's search space to find the suitable architecture. A \gls*{cnn} architecture search model contains many search spaces which are optimised in the image classification application. Google Brain has launched a search space for the NASNet model. It is a feed-forward network in which layers are broken into subgroups called cells. The normal cell can learn from images, and the normal cell maintains an image output equal to the input size. However, the kernel stride in each reduction cell is 2, halving the input image size. Normal and reduction cells are linked. Each cell is stacked during modelling, where $N$ normal cells are connected. Reduction cells are added between the $N$ normal cells, as shown in Fig.~\ref{fig:1} (left), to halve the image size, helping the next normal cell to process faster.

The output from the search space thereby includes normal cells and reduction cells used in the evaluation. \gls*{nas} has directed acyclic graphs (DAGs) connection between input $X1$ and $X2$ of cells, as shown in Fig.~\ref{fig:1} (right). In each normal cell, there are two input activations and one output activation. In the first normal cell, input $X1$ and $X2$ are copied from the input (image). The next normal cell uses, as input, both the $X1$ from the last normal cell, and the $X2$ from the second to last normal cell. All cells are connected the same way until the end of the model. Also, each cell has a greater number of cell output channels in each layer. 

Step 2: Evaluate the \gls*{cnn} model on a standard dataset for benchmarks. These benchmarks include number of errors, number of parameters, and search cost. The normal cells and reduction cells that are found in the search space are evaluated to measure the error rate, the number of parameters, and the computing costs, using CIFAR-10. Due to limited processor resources and GPU memory, parameters such as cell count ($N$), number of epochs, initial channel, and channel increment, are different for each search space and evaluation.

Step 3: Evaluate with a large-scale dataset.
When the model from the search space has been identified, it is evaluated with a larger dataset. Model evaluation with the CIFAR-10 dataset cannot be compared with other models because the CIFAR-10 dataset contains only ten classes. Given this constraint, CIFAR-100 datasets with 100 classes are required.

The search space of NASNet used \gls*{rl} and was tested with the CIFAR-10 dataset, which takes up to 2,000 GPU days to model. The AmoebaNet \cite{real2019regularized}, based on an evolutionary algorithm model, takes up to 3,150 GPU days for the same dataset. Also, the search space of NASNet was designed to use shorter search times. However, the sequential model-based optimisation (SMBO) method \cite{liu2018progressive} takes 335 GPU days, the gradient descent method \cite{liu2018darts} takes just 4 GPU days, whereas weight-sharing across different structures \cite{pham2018efficient} takes only 0.5 GPU days. 

As indicated, the AmoebaNet takes 3,150 GPU days, whereas the NSGA-Net \cite{lu2019nsga}, which uses a multi-objective evolutionary algorithm to find models, takes 4 GPU days. However, although the error rate of NSGA-Net is higher than that of AmoebaNet, based on a standard CIFAR-10 evaluation, the main focus of this area of research has been the reduction of search costs.

\subsection{Multi-objective Network Architecture Search}

\gls*{nas} aims to minimise errors, hyper-parameters, \gls*{flops}, and delays, making it challenging to identify a network architecture suitable for each objective simultaneously. Thus, the best network architecture should reduce or minimise all of these dimensions. For the evolution-based NASs, NSGA-Nets \cite{lu2019nsga} considers \gls*{flops} and error count, CARS \cite{yang2020cars} and LEMONADE \cite{elsken2018efficient} consider device-agnostic and device-aware objectives. In our work, however, we sought the achievement of the three goals; minimising errors, and reducing parameters and \gls*{flops}, simultaneously.

The NASs mostly focus on creating a network architecture for image recognition, then transferring that architecture to other tasks. However, for object detection and semantic segmentation, the same network architecture can be used as a backbone. 

Many network architectures, such as the EfficientNet \cite{tan2019efficientnet}, FBNetV2 \cite{wan2020fbnetv2}, DARTS \cite{liu2018darts}, P-DARTS \cite{chen2019progressive}, CDARTS \cite{yu2020cyclic}, CARS \cite{yang2020cars}, LEMONADE \cite{elsken2018efficient}, NSGA-Net \cite{lu2019nsga} and NSGA-NetV2 \cite{lu2020nsganetv2} were tested only on image recognition datasets. It is challenging to design and evaluate a network architecture for general purposes. 

Table~\ref{tab:1} shows the research objectives of the various NASs, illustrating that the image identification architectures were, in some cases, transferred to object detection, with one, MobileNetV3 \cite{howard2019searching} also being applied to transfer specifically researched image identification architecture to both object detection and semantic segmentation. 

Our objective was to extend this image identification architecture, using the ImageNet dataset, to object detection, semantic segmentation, as well the further purpose of keypoint detection.

\begin{table*}[]
\centering
\resizebox{\textwidth}{!}{%
\begin{threeparttable}
\begin{tabular}{lcccc}
\hline
\multicolumn{1}{c}{Methods} & Search Method & Multiple Objective & Dataset Searched   & Architecture transfer$\dagger$ \\ \hline
MobileNetV3 \cite{howard2019searching}  & RL + expert  & -   & ImageNet                        & IR, OD, SS \\
EfficientNet \cite{tan2019efficientnet} & RL + scaling & -   & ImageNet                        & IR         \\
FBNetV2 \cite{wan2020fbnetv2}           & gradient     & -   & ImageNet                        & IR         \\
DARTS \cite{liu2018darts}        & gradient     & -   & CIFAR-10                        & IR         \\
P-DARTS \cite{chen2019progressive}      & gradient     & -   & CIFAR-10, CIFAR-100             & IR         \\
PC-DARTS \cite{xu2020pc}     & gradient     & -   & CIFAR-10, ImageNet              & IR, OD     \\
CDARTS \cite{yu2020cyclic}      & gradient     & -   & CIFAR-10, ImageNet              & IR         \\
CARS \cite{yang2020cars}         & EA           & Yes & CIFAR-10                        & IR         \\
LEMONADE \cite{elsken2018efficient}    & EA           & Yes & CIFAR-10, CIFAR-100, ImageNet64 & IR         \\
NSGA-Net \cite{lu2019nsga}     & EA           & Yes & CIFAR-10                        & IR         \\
NSGA-NetV2 \cite{lu2020nsganetv2}   & EA           & Yes & ImageNet                        & IR         \\
EEEA-Net (this paper)       & EA+ EE-PI     & Yes                & CIFAR-10, ImageNet & IR, OD, SS, KD        \\ \hline
\end{tabular}

\begin{tablenotes}
    \item[$\dagger$] IR = Image Recognition, OD = Object Detection, SS = Semantic Segmentation, KD = Keypoint Detection.
\end{tablenotes}
\end{threeparttable}

}

\caption{Comparison of different NAS search method with multi-objectives.}
\label{tab:1}
\end{table*}

\subsection{Inverted Residuals Network (IRN)}

\begin{figure}[t]
\centering\includegraphics[width=0.90\linewidth]{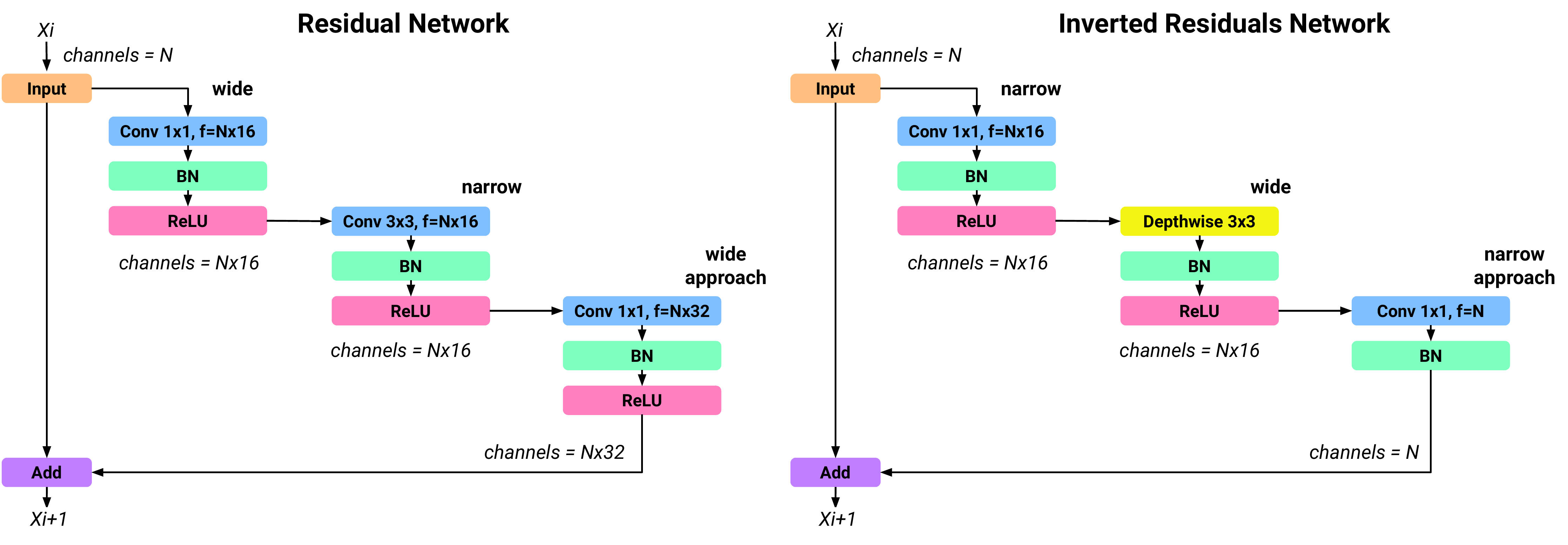}
\caption{The difference between the residual network (left) and the inverted residual network (right).}
\label{fig:4}
\end{figure}

The \gls*{irn} \cite{tan2019mnasnet} concept is needed to reduce the \gls*{rn} parameters. In contrast, the \gls*{rn} concept integrates data from the previous layer into the last layer. Fig.~\ref{fig:4} (left) shows that the \gls*{rn} structure has three layers: wide, narrow, and wide approach layers. The wide layers have $N\times16$ output channels whereas the narrow layers have $N\times16$ channels each. The wide approach layer has $N\times32$ output channels ($N$ is the input channels in each case). However, all the convolution layers used the standard convolution. Batch normalisation (BN) and activation functions (ReLU) were also added into each convolution layer. 
 
The \gls*{rn} structure is modified and reversed to obtain the \gls*{irn}. The layers in \gls*{irn} are defined as narrow layer, wide layer and narrow approach layer. In \gls*{irn}, the number of output channels obtained is equal to the number of input channels, $N$, as shown in Fig.~\ref{fig:4} (right). When the data is fed into the $1\times1$ convolution layer, the number of channels will be expanded to $N\times16$. The wide layer changes to a $3\times3$ depth-wise separable convolution instead of a $3\times3$ standard convolution, reducing the \gls*{flops} and number of parameters. There are $N\times16$ channels in a wide layer, which is equal to the previous layer. Also, $1\times1$ standard convolution is used in the narrow approach to reduce the channels' size to be equal to the input channel $N$. Then, the input data ($Xi$) is combined with the \gls*{irn}'s output to get data ($Xi + 1$). 


Convolutional layers can be defined in various formats to find a model structure with \gls*{ea}. Convolutional layers are also called cells. The types of convolutional layers used in our experiment are presented in Table~\ref{tab:2}. 


\begin{table}[t]
\centering
\begin{tabular}{ccc}
\hline
kernel      & Type                 & Layer       \\ \hline
3×3         & max and average      & pooling     \\
3×3 and 5×5 & depth-wise separable & convolution \\
3×3 and 5×5 & dilated              & convolution \\
3×3 and 5×5 & inverted residuals   & convolution \\
-           & skip connection      & -           \\ \hline
\end{tabular}
\caption{Search space of EA-Net and EEEA-Net.}
\label{tab:2}
\end{table}

\section{Method}

The most commonly used methods to develop \gls*{nas} are \gls*{rl} and gradient descent algorithms. However, these algorithms possess limitations in solving multi-objective problems. \gls*{ea} automates the model search process, is easier to implement, and enables discovery of solutions while considering multiple objectives. 

A general description of an \gls*{ea}, including encoding, a presentation of the multi-objective genetic algorithm, and genetic operations used with \gls*{nas}, is provided in Section~3.1. The Early Exit Population Initialisation concept and method, its simple, yet effective application in \gls*{ea}, mitigating the complexity and parameters used in the previous models, while maintaining the same accuracy, are described in Section~3.2.

\subsection{Evolutionary Neural Architecture Search}

The \gls*{ga} is an algorithm based on Darwinian concepts of evolution. The \gls*{ga} is part of a random-based \gls*{ea} \cite{xie2017genetic,baldominos2017evolutionary,real2017large}. \gls*{ga}'s search-based solutions stem from the genetic selection of robust members that can survive. The population is integral to \gls*{ga} because \gls*{ga} solutions are like organisms that evolve after the environment. The most suitable solutions need to rely on genetic diversity. Thus, a greater number of genetically diverse populations enables more effective \gls*{ga} solutions.

The initial phase of \gls*{ga} creates the first population for candidate solutions. The population is determined by population size, which describes total solutions. Each solution is called an individual, where an individual consists of a chromosome. The chromosome is a mix of genes. In the initial population, it is possible to provide unique information about each gene to all other genes, at random. A fitness function computes the fitness value for each individual. \gls*{cnn}'s model structure is searched with the \gls*{nas} search space, defining error rate as a fitness function where fitness value represents dataset error value. As shown in Equation~\ref{eq:1}, the fitness value is calculated where n is the number of individuals.

\begin{equation}
\label{eq:1}
fitness(i) = f(x_{i}), i = 1,2,3,..,n
\end{equation}

Organisms consist of different phenotypes and genotypes. Appearances such as foreign features (such as eye colour) and internal features (such as blood types) are called phenotypes. Genes of different organisms, called genotypes, can be transferred from model to model by gene transfer. 

The \gls*{cnn} model's architecture is represented as a genotype in the \gls*{nas} search space. A subset of the \gls*{nas} search space includes normal cells and reduction cells. The cells are stacked in the complete architecture. Normal cells or reduction cells consist of connected layers such as convolution layers, average pooling, max pooling, and skip connection. A complete model must connect cells to create a genotype for training or testing.


\begin{figure}[t]
\centering\includegraphics[width=0.90\linewidth]{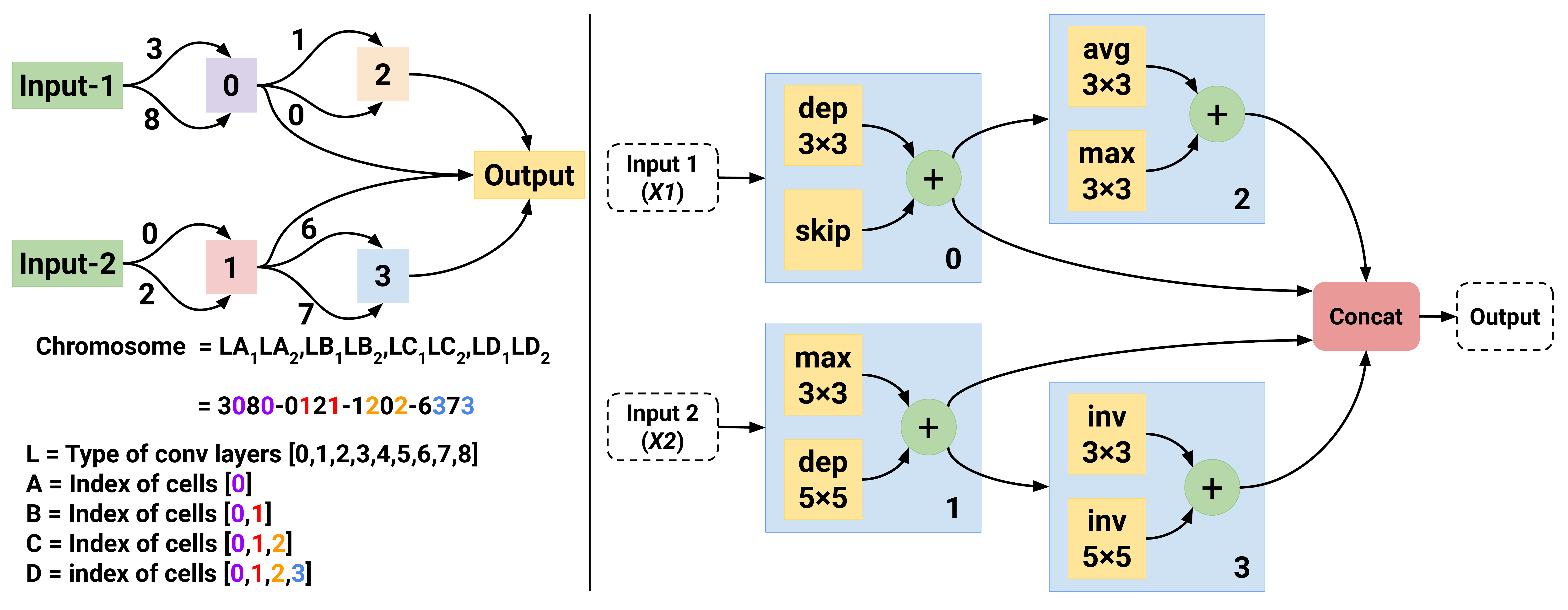}
\caption{The chromosome structure of the Evolutionary Algorithm.}
\label{fig:2}
\end{figure}

\subsubsection{Encoding}

The genotype of the \gls*{nas} model consists of normal cells and reduction cells, called chromosomes. There are various genes linked to chromosomes. The number of genes is defined as $LA_{1}LA_{2}$, $LB_{1}LB_{2}$, $LC_{1}LC_{2}$, and $LD_{1}LD_{2}$ as in Equation~\ref{eq:2}. The gene consists of operations ($L$) and indices of operations ($A, B, C, D$). Operation ($L$) can be considered a type of \gls*{cnn} layer, such as max pooling, average pooling, depth-wise separable convolution, dilated convolution, inverted residuals block, and skip connection. 


\begin{equation}
\label{eq:2}
chromosome(x) = LA_{1}LA_{2},LB_{1}LB_{2},LC_{1}LC_{2},LD_{1}LD_{2}
\end{equation}

For example, consider nine different operations ($L$) in the experiment, as $[0, 1, 2, 3, 4, 5, 6, 7, 8]$. Moreover, the operation index ($A, B, C, D$) refers to the operation location to be connected with other operations ($L$). From Fig.\ref{fig:2} (left), the index is defined as follows: A = $[0]$, B = $[0, 1]$, C = $[0, 1, 2]$, and D = $[0, 1, 2, 3]$. The connection between operations (L) and the operation index ($A, B, C, D$) determines the location of the connection between operations. For example, LA's gene code, $LA_{1}LA_{2}$ is 30,80, meaning output data processed in operation 3 and 8 will be linked to index 0.

Similarly, $LB_{1}LB_{2}$ equals 01,21, meaning data processed by operation 0 and 1 are connected at index 1. However, in the genes of $LC_{1}LC_{2}$ and $LD_{1}LD_{2}$, those genes are linked sequentially to the output of $LA_{1}LA_{2}$ and $LB_{1}LB_{2}$, to help reduce the number of model parameters. If the $LC_{1}LC_{2}$ and $LD_{1}LD_{2}$ is connected to the same input as $LA_{1}LA_{2}$, and $LB_{1}LB_{2}$ (parallel network) will increase the processing time and the parameters.

\begin{equation}
\label{eq:3}
Previous Index = index - 2
\end{equation}

The position of the previous index can be computed from Equation~\ref{eq:3}, where the index is greater than 1. If the index is an even number, it is linked to the even previous index; otherwise, it is linked to the odd previous index. Thus, the $LC_{1}LC_{2}$ gene is 12-02, which has an index of 2, and the $LC_{1}LC_{2}$ gene is linked from index 0. While the $LD_{1}LD_{2}$ gene is 63-73, it has an index of 3. The $LD_{1}LD_{2}$ gene is linked from index 1. However, if there are different indices in a gene, for example, a gene 63-72, operator 6 is connected from index 1 and operator 7 is from index 0.

\subsubsection{Multi-objective Genetic Algorithm}

Initially, \gls*{ga} was used for single-objective optimisation problems (SOOP) and, later, \gls*{ga} was developed to solve the multi-objective optimisation problem (MOOP) \cite{deb2002a}, which has more than one objective function to minimise fitness values. The \gls*{ga} that can solve the MOOP problems \cite{carrau2017enhancing,hasan2019dynamic} is called a multi-objective genetic algorithm (MOGA).


\begin{equation}
\label{eq:4}
\begin{aligned}
\min_{} \quad & \{f_{1} \left ( x \right ), f_{2} \left ( x \right ), ... , f_{k} \left ( x \right )\}\\
\textrm{s.t.} \quad & x\in X
\end{aligned}
\end{equation}

The optimisation of the \gls*{cnn} model is generally a problem with more than one objective. As illustrated in Equation~\ref{eq:4}, where $f$ is fitness values, the integer $k \ge 2$ is the number of objectives, $x$ is individual, and $X$ is the set of individuals. All these objectives must be as small as possible.  

Indicators used to measure CNNs model performance include model accuracy, model size, and processing speed. There are three objectives to consider during a model search: lowest validation error, minimum parameters, and computational cost. 


\begin{equation}
\label{eq:5}
\begin{aligned}
\min_{} \quad & \{ Error \left ( x \right ), FLOPS \left ( x \right ), Params \left ( x \right ) \}\\
\textrm{s.t.} \quad & x\in X \\
\quad & w_{error} + w_{flops}  + w_{params} = 1 \\
\quad & w_{error},w_{flops}, w_{params}  >=0
\end{aligned}
\end{equation}

The evolutionary algorithm finds the most effective model for each objective by finding the lowest objective values of the entire population. We defined the three objective values as being equally important. Thus, it is necessary to set the weight of each of the three objective values to 1/3 to find the best model for each value. As illustrated in Equation~\ref{eq:5}, where $x$ is individual, $X$ is the individual's set and $w_{error},w_{flops}, w_{params}$ weighs each objective's weight.

For the MOOP problem, it is almost impossible to find one solution that provides the optimal value for each objective function. For each solution given by the MOOP, the best solution group is called nondominated or Pareto optimal because these solutions are compared using the Pareto Domination principle. Many solutions can be obtained from the search using MOOP. These solutions will be reviewed again to find the best solution within the searched solution group.

The best solution group should not dominate when compared to other solutions. For example, any solution $v$ overwhelming a better solution can be represented as $v \prec w$. If no solution $v$ is worse than solution $w$, then solutions v is better than $w$.

\subsubsection{Genetic Operations}

The processes used to create offspring in the new generation are called genetic operations. A new population must replace an ancestor group that cannot survive. The population can be created in two ways, by crossover or mutation.

Crossover is the creation of a new population by switching genes of different chromosomes from two populations. The genotype of the parent chromosomes will be recombined to create a novel chromosome, which can be done in various ways. For example, a point crossover that performs random cutting points or chromosomes to produce offspring is a crossover between two chromosomes with a random probability of 0.5. Crossover creates offspring using random genes from parents.


\begin{figure}[t]
\centering\includegraphics[width=0.90\linewidth]{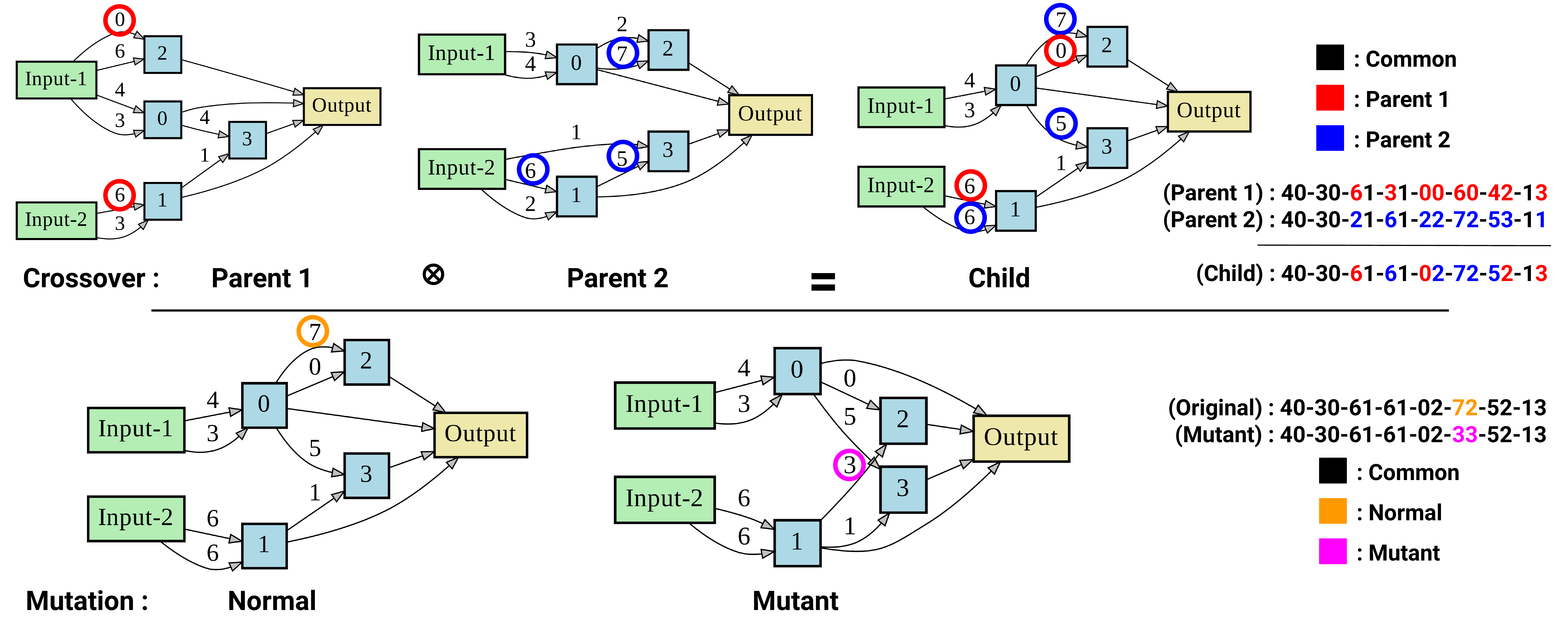}
\caption{Crossover operation (top): the parent has two different network architectures; the chromosome of each parent architecture can be visualised as a digits string. Child architecture was mixed with a chromosome index from their parent's chromosome. Mutation operation (bottom): The location of a normal chromosome was randomly selected. Then this pair of genes will be replaced by a new random pair.}
\label{fig:3}
\end{figure}

Fig.~\ref{fig:3} (top) demonstrates the uniform crossover operation used in this implementation, requiring two-parent architectures. We visualised the architectures as follows: 40-30-61-31-00-60-42-13, as parent 1 and 40-30-21-61-22-72-53-11, as parent 2. Then, in the crossover operation, the random probability of 0.5 is defined. The fifty-fifty chance was used to cross the gene between the two parent architectures for child modelling (40-30-61-61-02-72-52-13). The common parent gene is coloured black, but if the gene derived from the first parent is red, then the gene derived from the second parent is represented by blue. 

A mutation is an operation to reduce population uniformity and contribute to genetic diversity. The mutation changes data in the gene by randomly locating the gene and replacing the original gene with random new genes. The mutation causes offspring chromosomes to be different from parents. The individual being mutated is called a mutant.

Fig.~\ref{fig:3} (bottom) shows the mutation operation used during implementation; the location of a chromosome of architecture was determined by randomly selecting only one pair of gene locations (72, orange). Then it was replaced with a random pair of gene value (33, magenta) of the newly mutated gene.


\begin{figure*}[b!]
\centering\includegraphics[width=0.90\linewidth]{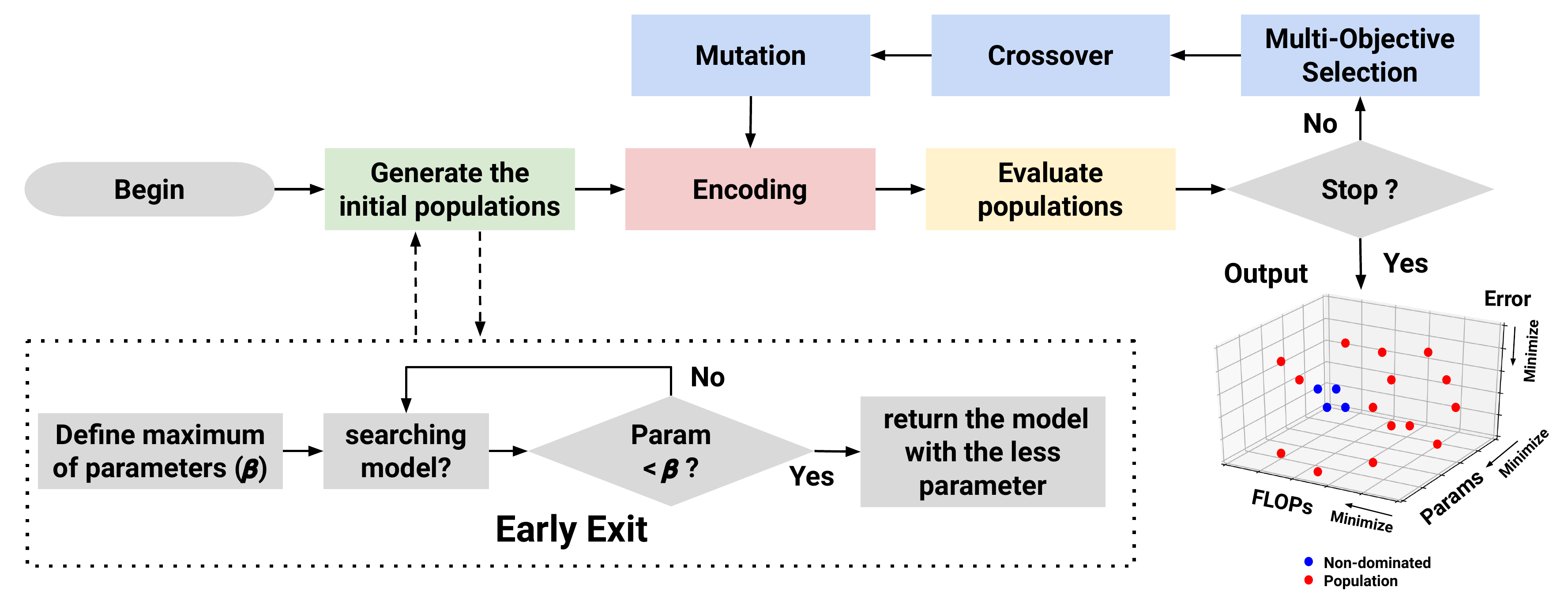}
\caption{An Early Exit Evolutionary Algorithm.}
\label{fig:5}
\end{figure*}

\subsection{Early Exit Population Initialisation (EE-PI)}

\gls*{ea} can find cell patterns by selecting the best model with the lowest error rate. However, the discovery process takes longer to search and select cells in each generation. Each population has to be trained and evaluated, which increases the time needed to find cell patterns.  
The single-objective \gls*{ea} uses the error value to find network architecture. However, the network architecture with the lowest error may have too many parameters. In our experiment, the maximum number of generations was set to 30 with 40 populations per generation due to the limited resources of a single GPU.

The population is the only factor affecting processing time. The evaluation must examine the entire population to select the population with the lowest error rate, thus obtaining an effective model structure. However, a longer search time is required to evaluate every population using a single processor unit. Therefore, the \gls*{eepi} method was introduced into the evolutionary algorithm to reduce the search time and control the number of parameters, as illustrated in Fig.~\ref{fig:5} and detailed in Algorithm~\ref{alg:1}. 

\begin{algorithm}[t!]
	\caption{Multi-objective evolutionary algorithm with an Early Exit Population Initialisation.}
	\label{alg:1}
	\begin{algorithmic}
		\STATE {\bfseries Input:} The number of generations $G$, population size $n$, validation dataset $D$, objectives $w$.
		\STATE {\bfseries Output: } A set of $K$ individuals on the Pareto front.
		\STATE {\bfseries Initialisation: } An Early Exit Population Initialisation $P_{1}$ and $Q_{1}$.

		\FOR {$i=1$ {\bfseries to} $G$}
		\STATE{$R_i = P_i \cup Q_i$}
		
		\FORALL { $p \in R_{i}$ }
		\STATE{Train model $p$ on $D$}
		\STATE{Evaluate model $p$ on $D$}
		\ENDFOR
		
		\STATE {$ M = \text{tournament-selection}(P_{i+1})$}
		\STATE {$F = \text{non-dominated-sorting}(R_i)$}
		
		\STATE{Pick $n$ individuals to form $P_{i+1}$ by ranks and the crowding distance \textbf{weighted} by $w$ based on Equation~\ref{eq:5}}

		\STATE {$Q_{i+1} = \text{crossover}(M) \cup \text{mutation}(M)  $ }

		\ENDFOR
		
		\STATE {Select $K$ models at an equal distance near Pareto front from $P_{G}$}
	\end{algorithmic}
\end{algorithm}

The \gls*{eepi} method filters the \gls*{cnn} models based on the number of parameters in the network architecture, which is iteratively compared to a pre-set maximum value ($\beta$). The \gls*{eepi} obtains the \gls*{cnn} network architecture which has less parameters than the maximum number of parameters attached to the \gls*{ea}, as illustrated in Fig.~\ref{fig:5}, which shows the Early Exit as the dashed-line block.

\begin{equation}
\label{eq:6}
\text{EarlyExit}({\alpha}, {\beta})\triangleq \left\{\begin{matrix}
1, & \text{if} \ {\alpha}\le {\beta} \\ 0, & \text{otherwise}
\end{matrix}\right. 
\end{equation}

In Equation~\ref{eq:6}, where $\alpha$ is the parameter of the model which is discovered, $\beta$ is the maximum number of parameters. If the number of parameters found in the model are less than or equal to the maximum number of parameters (${\alpha}\le {\beta}$), then the model will be considered as a part of the first-generation population.

For example, to select a network architecture with a maximum of 3 million parameters ($\beta = 3$), \gls*{ea} selects the model by considering the number of parameters lower than the maximum number of parameters. Suppose network architecture is not considered, because it has more than maximum parameters ($\beta$). In this case, it chooses a new structure with less than 3 million parameters. Therefore, in conjunction with the \gls*{ea} in the selection process, Early Exit facilitates the filtering out of the model with the number of parameters greater than the maximum number of parameters. The best network architecture is also discovered using the \gls*{ea} with Early Exit by considering the error rate and the number of parameters.

\section{Experiments and Results}

Experiments were carried out in three parts: First, finding and evaluating the network architecture with EEEA on CIFAR-10 and CIFAR-100. The second part was the finding and evaluation of the EEEA-Net using the ImageNet datasets. In the third part, the EEEA-Net obtained from the second part is applied for other tasks such as object detection, semantic segmentation, and keypoint detection. The PyTorch deep learning library was used in the experiments. The experiment was carried out on Intel(R) Xeon(R) W-3235 CPU @ 3.30GHz 12 Core CPU, 192 GB RAM and NVIDIA RTX 2080 Ti GPU, running on the Ubuntu 18.04.3 operating systems.

\subsection{CIFAR-10 and CIFAR-100 datasets}
This subsection searched for a model with the CIFAR-10 dataset; it was evaluated on CIFAR-10 and CIFAR-100 datasets. Both CIFAR-10 and CIFAR-100 datasets consisted of 60,000 images, with  50,000 images and 10,000 images in the training set and test set, respectively. The CIFAR-10 and CIFAR-100 have 10 and 100 classes, respectively, with 600 images in each class.

\subsubsection{Architecture Search on CIFAR-10}

Thirty generations with 40 populations in each generation were defined to locate the network architecture with \gls*{ea}. The first-generation populations were randomly generated, with subsequent populations in generations 2-30 being evolved with \gls*{ea}. Each population was defined with a depth of two normal cells instead of the usual six normal cells. Thus, it reduced the search time of the network architecture. The search and evolution process happens more rapidly when Early Exit is used in the initial populations' process. Early Exit selects the network architecture having less than the pre-specified maximum of parameters ($\beta$). Thus, population evolution will choose only network architectures that are efficient and have fewer parameters. 

The hyper-parameters for the search process were defined as: the total number of cells (normal cells and reduce cells) equal to eight layers with 32 initial channels by training the network from scratch for one epoch on the CIFAR-10 dataset. The hyper-parameters used included a batch size of 128, with SGD optimiser with weight decay equal to 0.0003 and momentum equal to 0.9. The initial learning rate was 0.05. Using the cosine rule scheduler, the Cutout regularisation had a length set to 16, a drop-path of the probability of 0.2, and the maximum number of parameters equal to 3, 4, and 5 million. 

The evolutionary algorithm (EA-Net, $\beta=0$) took 0.57 GPU days to find the network architecture with NVIDIA RTX 2080 Ti. However, the early exit evolutionary algorithm (EEEA-Net-A, $\beta=3$) took 0.38 GPU days, EEEA-Net-B ($\beta=4$) took up to 0.36 GPU days, and EEEA-Net-C ($\beta=5$) took up to 0.52 GPU days. These architectures are used for performance evaluation in the next section. 

\subsubsection{Architecture Evaluation on the CIFAR-10 dataset}

The network architecture had to be changed to find the normal and reduced cells with the Early Exit evolutionary algorithms. The CIFAR-10 dataset was used for the evaluation. The hyper-parameters were defined with the number of all cells (normal and reduce cells) set to 20 layers with 32 initial channels, the network was trained from scratch with 600 epochs with a batch size of 96, SGD optimiser with weight decay was 0.0003 and momentum 0.9, and the initial learning rate set to 0.025. Using the cosine rule scheduler, the Cutout regularisation had a length set to 16, a drop-path of the probability of 0.2, and auxiliary towers of weight equal to 0.4.

\begin{table}[]
\centering{%
\begin{tabular}{lccccc}
\hline
\multicolumn{1}{c}{Architecture} &
  \begin{tabular}[c]{@{}c@{}}CIFAR-10 Error\\ (\%)\end{tabular} &
  \begin{tabular}[c]{@{}c@{}}CIFAR-100 Error\\ (\%)\end{tabular} &
  \begin{tabular}[c]{@{}c@{}}Params\\ (M)\end{tabular} &
  \begin{tabular}[c]{@{}c@{}}Search cost\\ (GPU days)\end{tabular} &
  \begin{tabular}[c]{@{}c@{}}Search\\ Method\end{tabular} \\ \hline
NASNet-A + CO \cite{zoph2016neural}             & 2.83          & 16.58          & 3.1          & 2,000        & RL        \\
ENAS + CO \cite{pham2018efficient}                & 2.89          & 17.27          & 4.6          & 0.5          & RL        \\ \hline
PNAS \cite{liu2018progressive}                     & 3.41          & 17.63          & 3.2          & 225          & SMBO      \\ \hline
DARTS-V1 + CO \cite{liu2018darts}            & 2.94          & -              & 2.9          & 1.5          & gradient  \\
DARTS-V2 + CO \cite{liu2018darts}               & 2.83          & 17.54          & 3.4          & 4            & gradient  \\
P-DARTS + CO \cite{chen2019progressive}             & 2.50           & 15.92          & 3.4          & 0.3          & gradient  \\
PC-DARTS + CO \cite{xu2020pc}           & 2.57          & 17.36          & 3.6          & \textbf{0.1} & gradient  \\
CDARTS + CO \cite{yu2020cyclic}             & 2.48          & 15.69          & 3.8          & 0.3          & gradient  \\ \hline
AmoebaNet-A + CO \cite{real2019regularized}         & 3.12          & 18.93          & 3.1          & 3,150        & evolution \\
AmoebaNet-B + CO \cite{real2019regularized}         & 2.55          & -              & 2.8          & 3,150        & evolution \\
LEMONADE \cite{elsken2018efficient}                  & 3.05          & -              & 4.7          & 80           & evolution \\
NSGA-Net + CO \cite{lu2019nsga}            & 2.75          & 20.74          & 3.3          & 4            & evolution \\
CARS-I + CO \cite{yang2020cars}             & 2.62          & 16.00             & 3.6          & 0.4          & evolution \\
EA-Net ($\beta = 0$) + CO     & 3.30           & 17.58          & 2.9          & 0.57         & evolution \\
EEEA-Net-A ($\beta = 3$) + CO & 3.69          & 20.16          & \textbf{1.8} & 0.34         & evolution \\
EEEA-Net-B ($\beta = 4$)+ CO  & 2.88          & 16.90           & \textbf{1.8} & 0.36         & evolution \\
EEEA-Net-C ($\beta = 5$)+ CO  & \textbf{2.46} & \textbf{15.02} & 3.6          & 0.52         & evolution \\ \hline
\end{tabular}%
}
\caption{Comparing EEEA-Net with other architectures from RL, SMBO, gradient, and evolution search method on CIFAR-10 and CIFAR-100 datasets.}
\label{tab:3}
\end{table}

Table~\ref{tab:3} shows the comparisons and evaluations of EEEA-Net with other state-of-the-art models. EEEA-Net-C was evaluated with the test dataset, giving an error rate of 2.46\% for CIFAR-10. It took 0.52 GPU days to find normal and reduce cells. 

By comparison, our EEEA-Net-C model achieved a lower error rate and search time than all of those other models. 

AmoebaNet-B was the lowest of the other state-of-the-art models: NASNet-A model, PNAS, both DARTS versions, and NSGA-Net. However, the AmoebaNet-B model required 3,150 GPU days to complete, according to \cite{real2019regularized}. This is clearly a hugely greater amount of search resources than required in our model (0.52 GPU days).

\subsubsection{Performance and Computational Complexity Analysis}
The multi-objective search tested error rate, number of \gls*{flops}, and parameters. Optimisation on the effectiveness of a multi-objective uses Hypervolume (HV) as a measure of performance that computes the dominated area, using a reference point (Nadir point) with the most significant objective value from the first-generation population. Then, the Pareto-frontier solution computes the area between the reference point and Pareto. The higher HV shows that a multi-objective solution performs better in all objectives. 


\begin{figure}[t!]
\centering\includegraphics[width=0.75\linewidth]{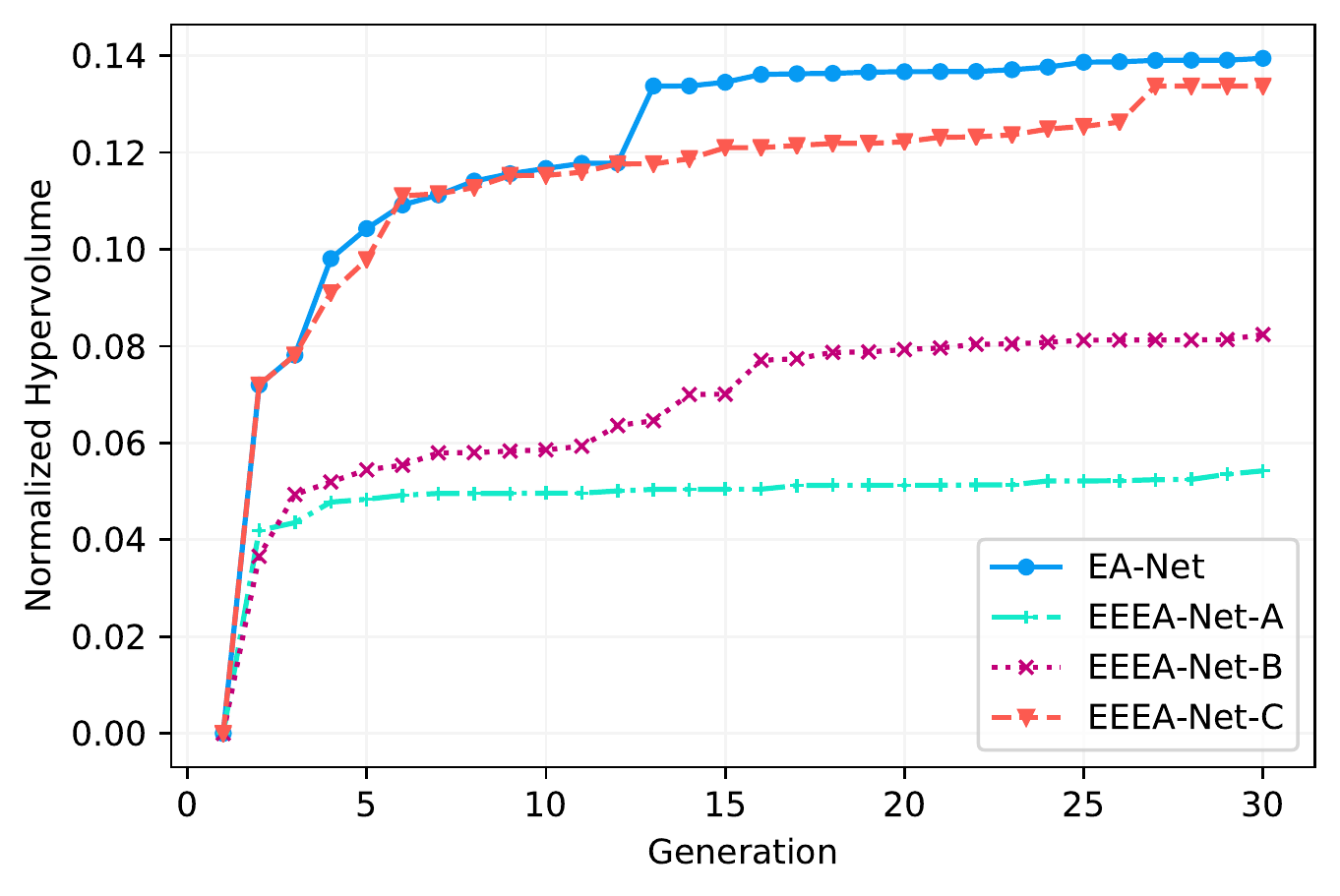}
\caption{Performance Metric of EA-Net and EEEA-Nets.}
\label{fig:6}
\end{figure}

After the model search, the HV for all the solutions obtained from the search is calculated to compare the performance of two variants: a model that uses Early Exit (EA-Net) and a model that does not use Early Exit. In Fig.~\ref{fig:6}, the values shown in the vertical axis are normalised HV, and the horizontal axis is generations. When we closely look at the HV value, it was found that the search using the EA-Net model yielded HV values greater than the three EEEA-Net models. 

However, considering only the model with an early exit, it was found that searches using $\beta$ equal to 5 performed better than $\beta$ equal to 3 and 4, since $\beta$ is a parameter that determines the model size by the number of parameters. Consequently, creating larger models by increasing the size of $\beta$, gave superior performance.

In addition, when considering a model without an Early Exit (EA-Net) and a model that used an Early Exit ($\beta$=5, EEEA-Net-C), it was found that the search efficiency of the EEEA-Net-C model was nearly similar to that of EA-Net because the EA-Net search does not control the model size while searching for the model. Therefore, the model obtained by EA-Net's search may is likely to obtain a model of a large parameter size. On the other hand, the model with an Early Exit better controls the model size, and the resulting model provides similar performance than achievable in an uncontrolled search.

\begin{figure*}[]
\centering

\subfloat[CIFAR-10 Accuracy vs Generations.]{\includegraphics[width=0.4\linewidth]{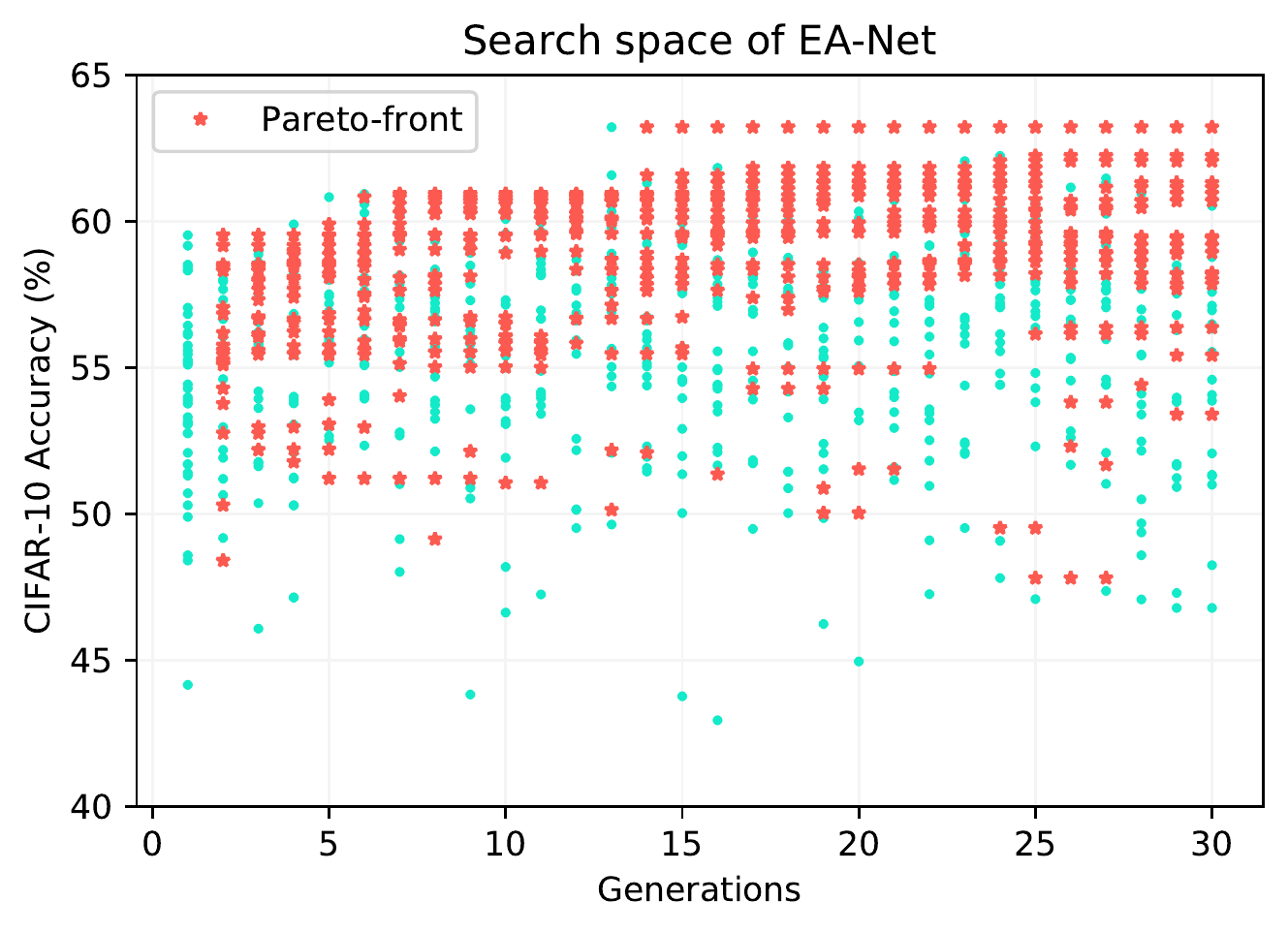}
\label{fig:7-1-1}}
\qquad
\subfloat[CIFAR-10 Accuracy vs FLOPS.]{\includegraphics[width=0.4\linewidth]{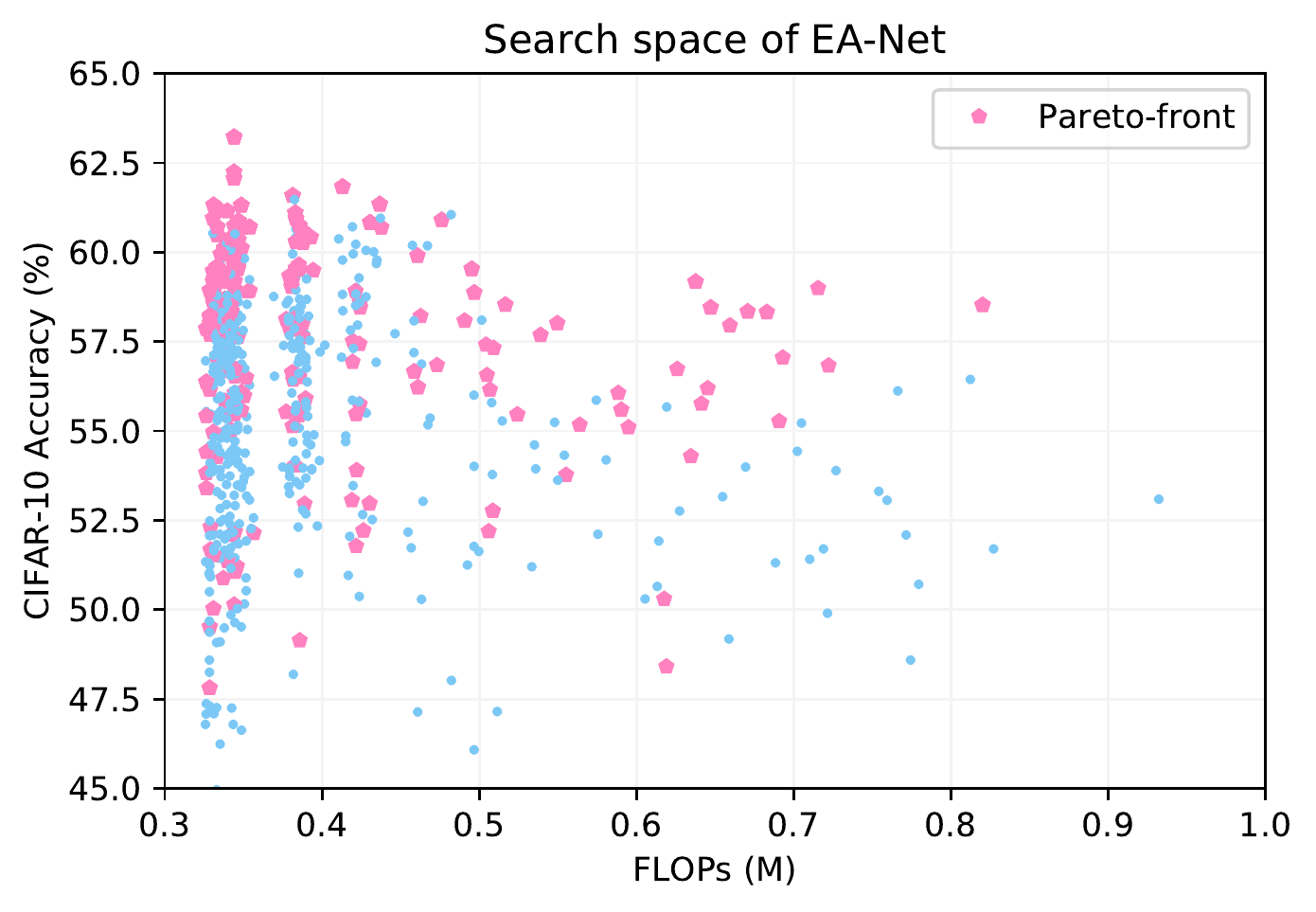}
\label{fig:7-2-1}}

\subfloat[CIFAR-10 Accuracy vs Generations.]{\includegraphics[width=0.4\linewidth]{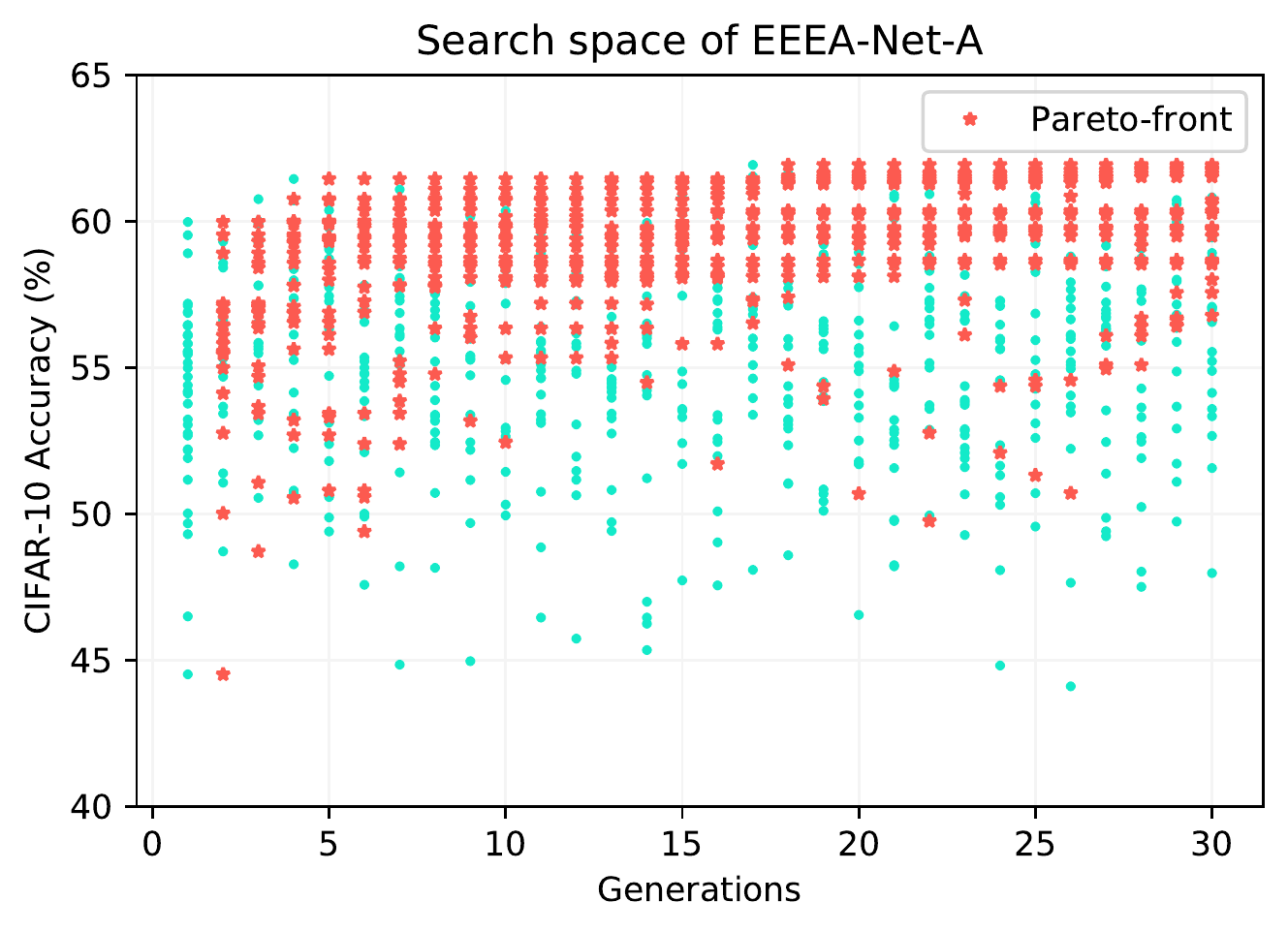}
\label{fig:7-1-2}}
\qquad
\subfloat[CIFAR-10 Accuracy vs FLOPS.]{\includegraphics[width=0.4\linewidth]{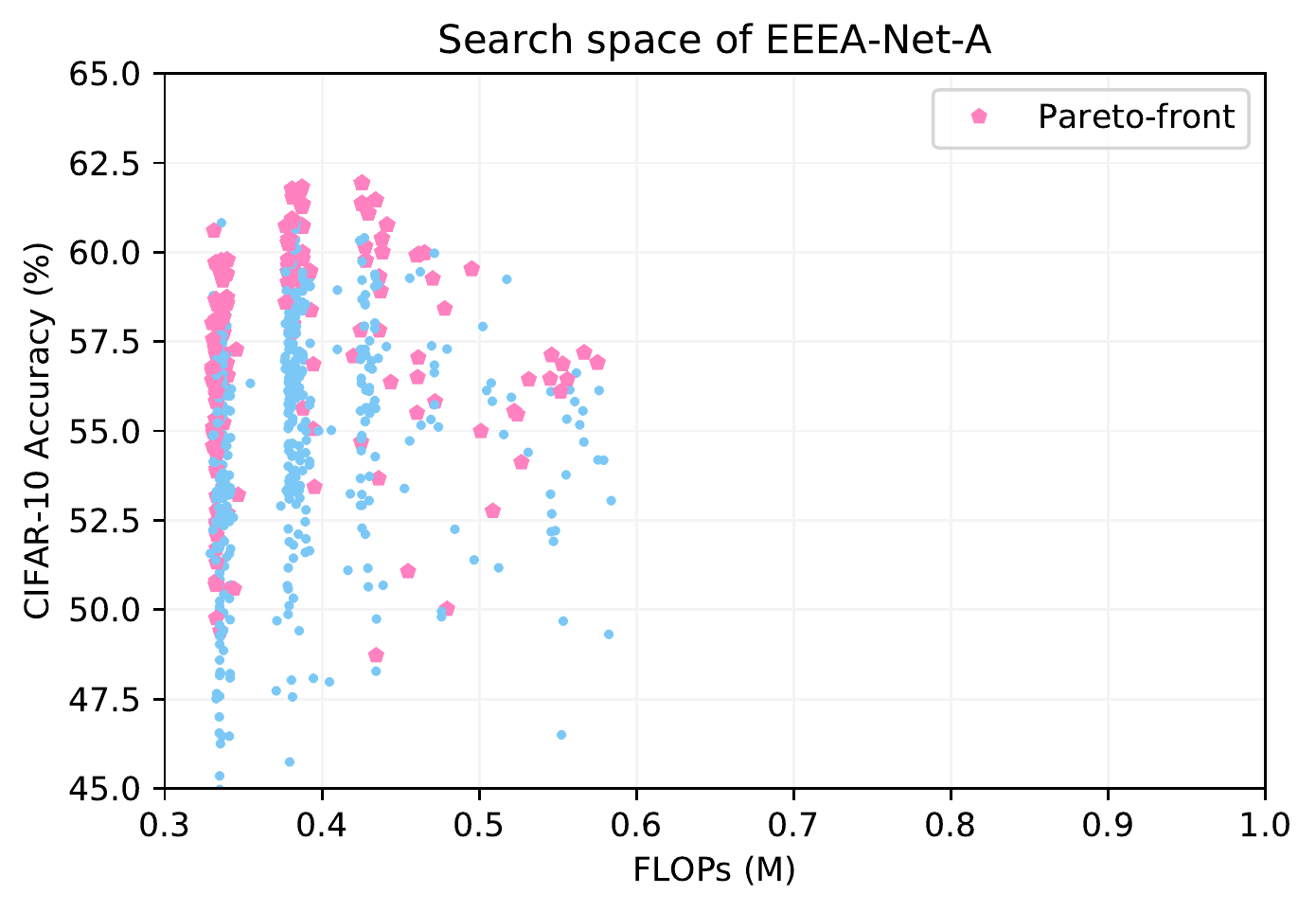}
\label{fig:7-2-2}}

\subfloat[CIFAR-10 Accuracy vs Generations.]{\includegraphics[width=0.4\linewidth]{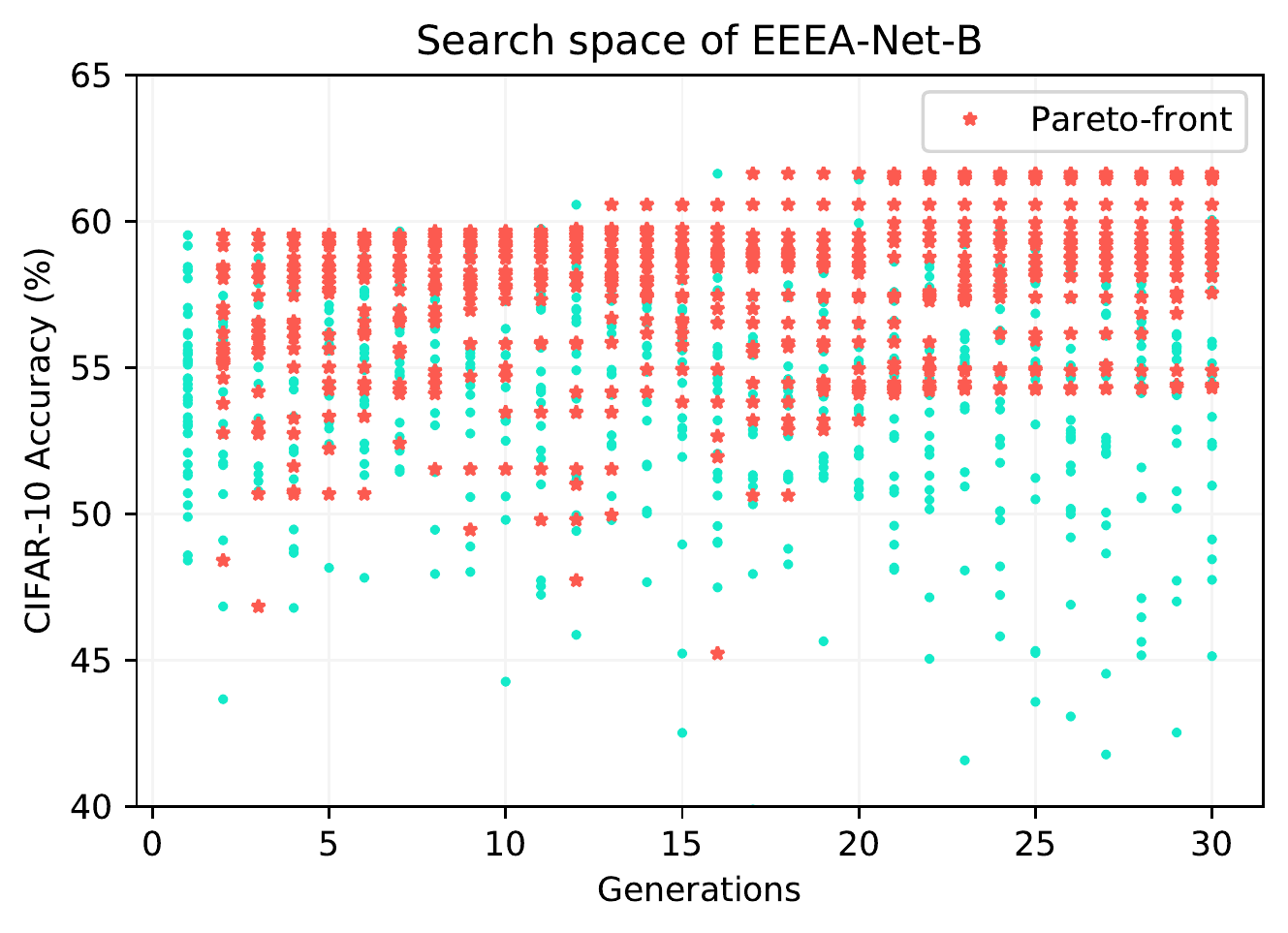}
\label{fig:7-1-3}}
\qquad
\subfloat[CIFAR-10 Accuracy vs FLOPS.]{\includegraphics[width=0.4\linewidth]{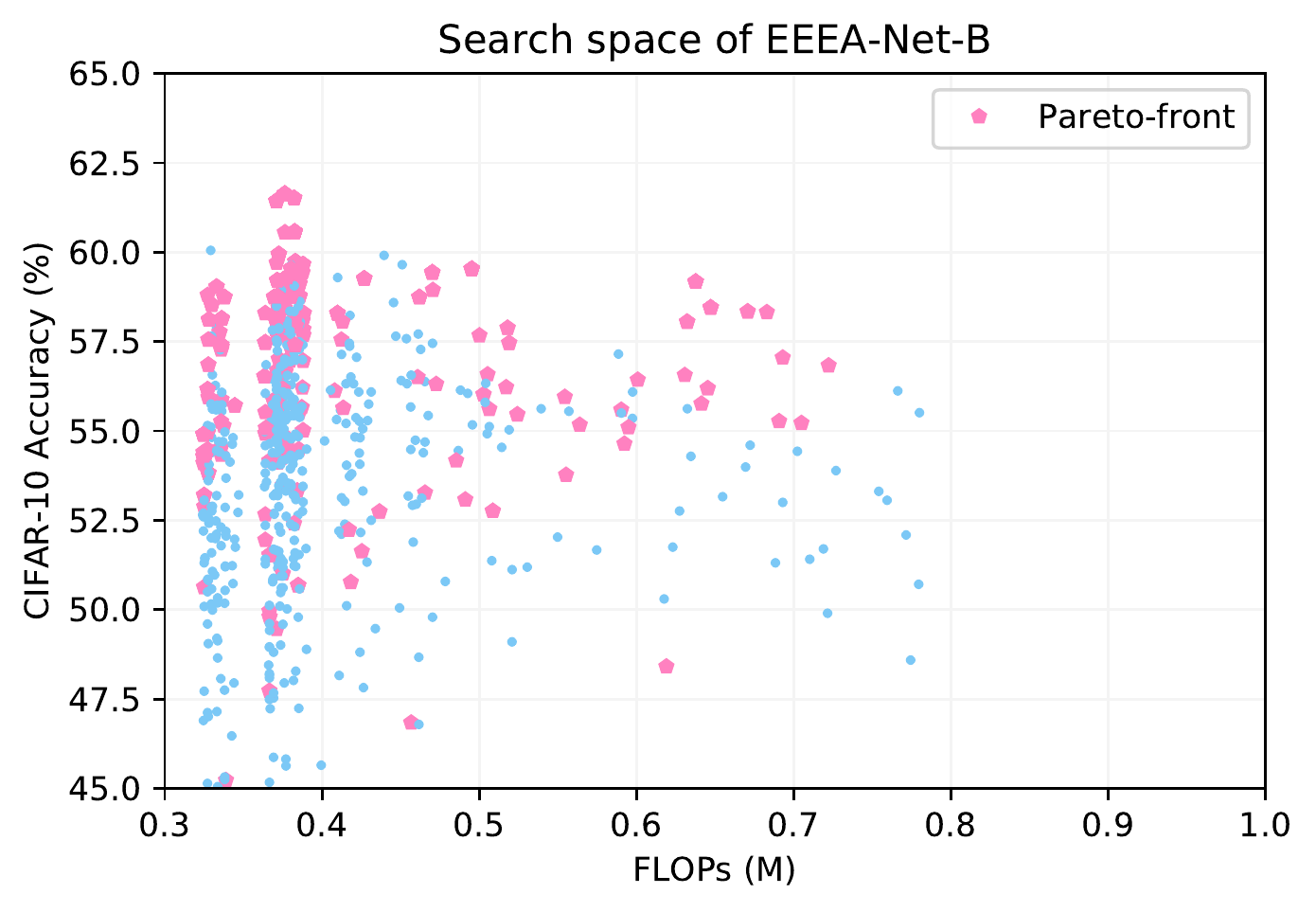}
\label{fig:7-2-3}}

\subfloat[CIFAR-10 Accuracy vs Generations.]{\includegraphics[width=0.4\linewidth]{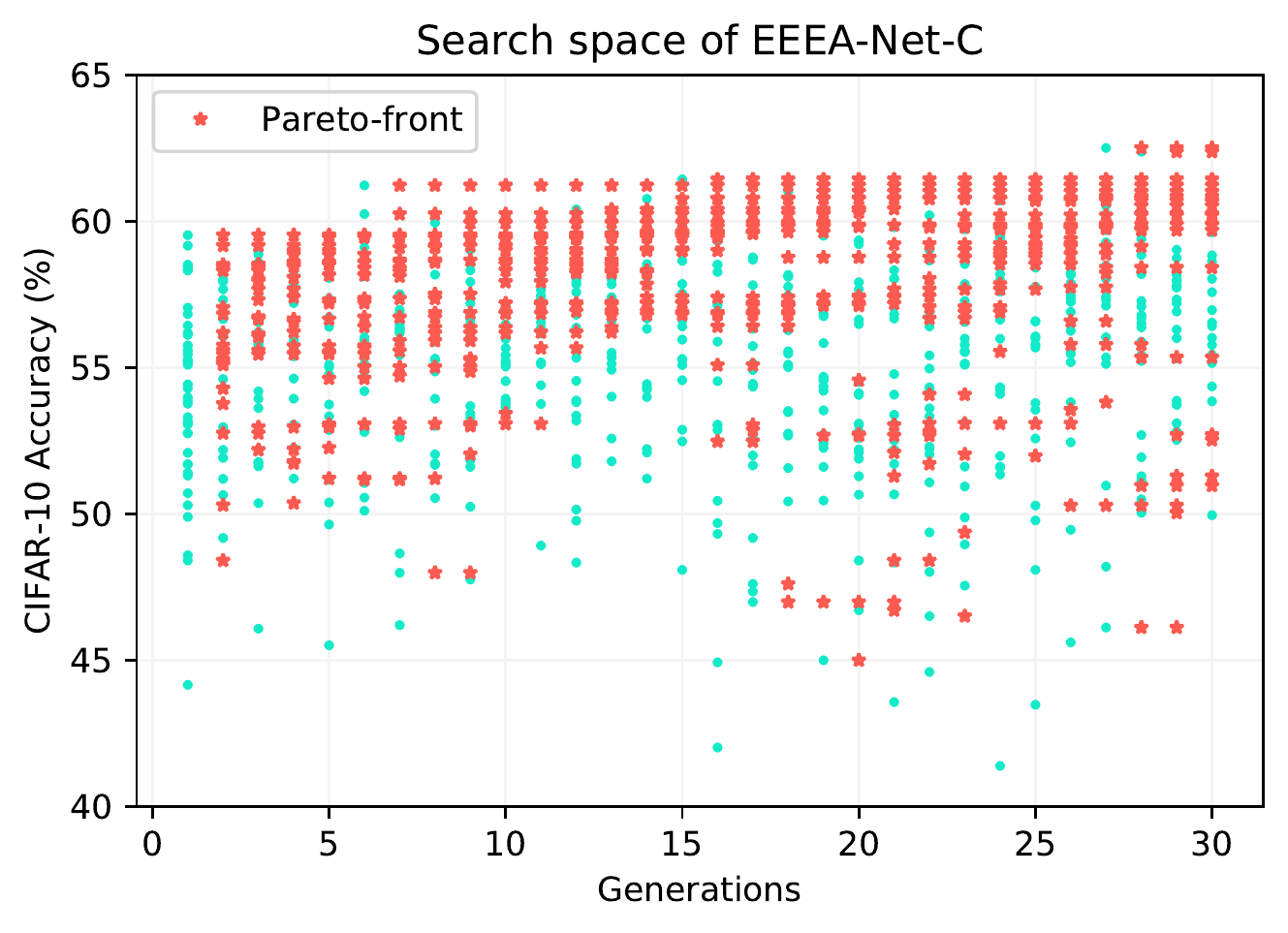}
\label{fig:7-1-4}}
\qquad
\subfloat[CIFAR-10 Accuracy vs FLOPS.]{\includegraphics[width=0.4\linewidth]{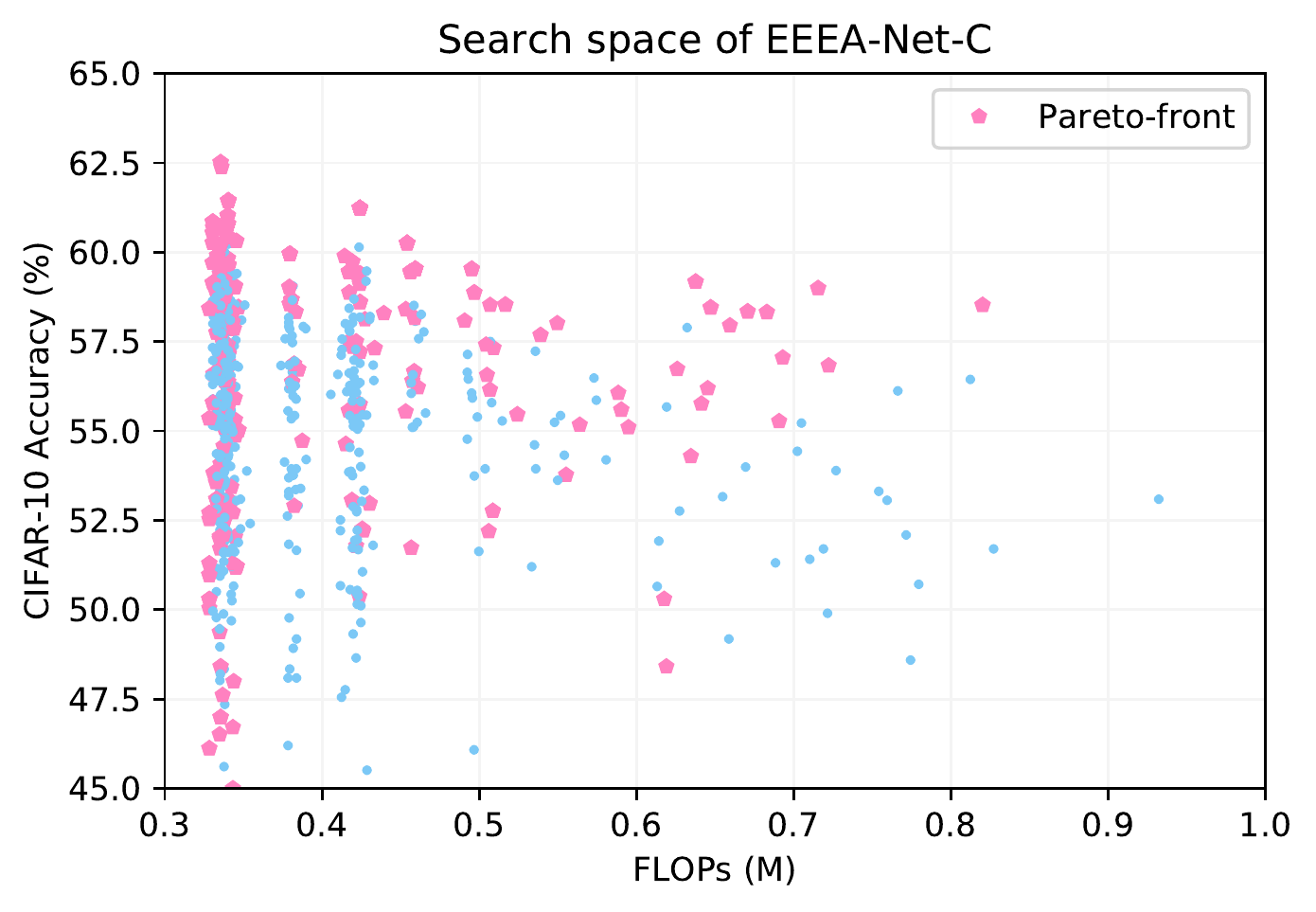}
\label{fig:7-2-4}}

\caption{Progress of trade-offs after each generation of EA-Net and EEEA-Nets.}
\label{fig:7}
\end{figure*}
 
We present progress trade-offs after each generation of the EA-Net and EEEA-Nets search through Fig.~\ref{fig:7}. The whole population is demonstrated by two-dimensional coordinates such as CIFAR-10 accuracy vs. Generations and CIFAR-10 accuracy vs. \gls*{flops}.

\subsubsection{Data augmentation}

The results from the evaluation of EEEA-Net are represented in precision floating-point (FP32). Our experimental goals were to create the most effective EEEA-Net without modifying the model structure. In the evaluation, we added the \gls*{aa} \cite{cubuk2019autoaugment} technique to the data augmentation process. \gls*{aa} created a more diverse set of data, making the model more effective. When Cutout \cite{devries2017improved} and \gls*{aa} \cite{cubuk2019autoaugment} were used, we observed that the error rate of EEEA-Net-C was reduced to 2.42\%. Without \gls*{aa}, however,  an error rate of 2.46\% occurred, as shown in Table~\ref{tab:4}.

\begin{table}[b!]
\centering
{%
\begin{tabular}{lccc}
\hline
\multicolumn{1}{c}{Architecture} &
  \begin{tabular}[c]{@{}c@{}}CIFAR-10 Error\\ (\%)\end{tabular} &
  \begin{tabular}[c]{@{}c@{}}Training time\\ (GPU Hours)\end{tabular} &
  AutoAugment \\ \hline
EEEA-Net-A ($\beta= 3$) + CO & 3.69          & 25.75 & -   \\
EEEA-Net-B ($\beta= 4$) + CO & 2.88          & 25.95 & -   \\
EEEA-Net-C ($\beta= 5$) + CO & \textbf{2.46} & 48.05 & -   \\ \hline
EEEA-Net-A ($\beta= 3$) + CO & 3.35          & 30.38 & Yes \\
EEEA-Net-B ($\beta= 4$) + CO & 2.87          & 31.26 & Yes \\
EEEA-Net-C ($\beta= 5$) + CO & \textbf{2.42} & 54.26 & Yes \\ \hline
\end{tabular}%
}
\caption{Results of CIFAR-10 using Cutout (CO) and AutoAugment (AA).}
\label{tab:4}
\end{table}

\subsubsection{Architecture Evaluation on CIFAR-100 dataset}

The \gls*{aa} technique was used to optimise the search and evaluation process of the EEEA-Net. The EEEA-Net-C was trained using the CIFAR-10 dataset, which was not sufficient for our purposes. Consequently, the EEEA-Net architectures obtained from CIFAR-10 dataset were used with CIFAR-100.  

The hyper-parameters used in the training process were changed to evaluate the EEEA-Net with the CIFAR-100 dataset, where the number of all cells (normal and reduce cells) was set to 20 layers with 36 initial channels. This was the outcome of training the network from scratch in 600 epochs with a batch size of 128, setting the SGD optimiser with a weight decay of 0.0003 and momentum of 0.9, and the initial learning rate set to 0.025 and running with the cosine rule scheduler. The Cutout regularisation length was equal to 16, and the drop-path of probability was 0.2, with auxiliary towers of the weight of 0.4. 

When the EEEA-Net-C (same model structure) was evaluated with CIFAR-100 datasets, it showed an error rate of 15.02\%, as shown in Table~\ref{tab:3}. Further, this evaluation, with 3.6 million parameters, resulted in the lowest error rate of all the state-of-the-art models. 

\subsection{ImageNet dataset}
In this subsection, we used the ImageNet dataset for the search and model evaluation. The ImageNet dataset is a large-scale standard dataset for benchmarking performance for image recognition for 1,000 classes with 1,281,167 images for the training set, 50,000 images for the test set, divided into 1,000 classes.

\begin{table}[b!]
\centering
\resizebox{\textwidth}{!}{%
\begin{tabular}{lccccc}
\hline
\multicolumn{1}{c}{Model} & Top-1 Error (\%) & Top-5 Error (\%) & Params   (M) & FLOPS (M) & Type \\ \hline
GhostNet 1.0 \cite{han2020ghostnet}               & 26.1          & 8.6          & 5.2          & 141          & manual   \\
MobileNetsV3 Large 0.75 \cite{howard2019searching}   & 26.7          & -            & \textbf{4.0} & 155          & combined \\
EA-Net-N1 (ours)                & 26.1          & 8.6          & 4.4          & 140          & auto     \\
EEEA-Net-A1 ($\beta = 5$)(our)  & 26.3          & 8.8          & 5.0          & \textbf{127} & auto     \\
EEEA-Net-B1 ($\beta = 6$) (our) & 26.0          & \textbf{8.5} & 5.0          & \textbf{138} & auto     \\
EEEA-Net-C1 ($\beta = 7$) (our) & \textbf{25.7} & \textbf{8.5} & 5.1          & 137          & auto     \\ \hline
MobileNetsV1 \cite{howard2017mobilenets}                & 29.4          & -            & 4.2          & 575          & manual   \\
MobileNetsV2 \cite{sandler2018mobilenetv2}             & 28.0          & -            & 3.4          & 300          & manual   \\
GhostNet 1.3 \cite{han2020ghostnet}               & 24.3          & 7.3          & 7.3          & 226          & manual   \\
MobileNetsV3 Large 1.0 \cite{howard2019searching}     & 24.8          & -            & 5.4          & 219          & combined \\
NASNet-A \cite{zoph2016neural}                    & 26.0          & 8.4          & 5.3          & 564          & auto     \\
MnasNet-A1 \cite{tan2019mnasnet}                 & 24.8          & 7.5          & \textbf{3.9} & 312          & auto     \\
FBNet-C \cite{wu2019fbnet}                    & 25.1          & -            & 5.5          & 375          & auto     \\
MOGA-A \cite{chu2020moga}                     & 24.1          & \textbf{7.2} & 5.1          & 304          & auto     \\
FairNAS-A \cite{chu2019fairnas}                  & 24.7          & 7.6          & 4.6          & 388          & auto     \\
PNASNet-5 \cite{liu2018progressive}                  & 25.8          & 8.1          & 5.1          & 588          & auto     \\
NSGANetV1-A2 \cite{lu2020multi}               & 25.5          & 8.0          & 4.1          & 466          & auto     \\
OnceForAll \cite{cai2020once}                 & 24.0          & -            & 6.1          & 230          & auto     \\
MSuNAS \cite{cai2020once}                     & 24.1          & -            & 6.1          & 225          & auto     \\
EA-Net-N2 (ours)                & 24.4          & 7.6          & 5.9          & 226          & auto     \\
EEEA-Net-A2 ($\beta = 5$) (our) & 24.1          & 7.4          & 5.6          & \textbf{198} & auto     \\
EEEA-Net-B2 ($\beta = 6$) (our) & 24.0          & 7.5          & 5.7          & \textbf{219} & auto     \\
EEEA-Net-C2 ($\beta = 7$) (our) & \textbf{23.8} & 7.3          & 6.0          & \textbf{217} & auto     \\ \hline
\end{tabular}%
}
\caption{Comparing EEEA-Net with other architectures from manual, combined, and auto search method on ImageNet datasets.}
\label{tab:5}
\end{table}

\subsubsection{Architecture Search on ImageNet}

Early Exit was used to discover a network architecture using the CIFAR-10 dataset. However, this network architecture was constructed from a multi-path \gls*{nas}, which requires considerable memory. Given this, we used a single-path \gls*{nas} to find the network architecture on ImageNet to reduce this search time, which also allows a multi-objective search with early exit population initialisation to be used on the OnceForAll \cite{cai2020once} super-network (called Supernet) to discover all network architectures that offer the best trade-off. Supernet can also search for the four dimensions of the network architecture, including kernel size, width (number of channels), depth (number of layers), and input resolution resize. We set all hyper-parameters for our architecture searches following the process in NSGA-NetV2 \cite{lu2020nsganetv2}.

The two objectives of accuracy and \gls*{flops} were the criteria for searching for 300 high accuracy samples with low \gls*{flops}. However, these sample architectures have a diverse number of parameters. The number of parameters affects architecture size when running on devices that may have memory constraints. Thus, to prevent the architecture from having too many parameters, we appended the Early Exit to create the first population with limited parameters. 

In this experiment, we compiled the number of architecture parameters shown in Table~\ref{tab:5} to calculate the average number of parameters equal to 5. Thus, the maximum number of parameters ($\beta$) where $\beta$ equals 5, 6 or 7, was defined as follows: EEEA-Net-A ($\beta=5$), EEEA-Net-B ($\beta=6$), EEEA-Net-C ($\beta=7$). For a fair comparison, we set $\beta$ qual to 0, and called that EA-Net-N ($\beta=0$). We categorised our networks using the number of MobilenetV3 \gls*{flops} to define the network size architectures of EEEA-Net-A1, EEEA-Net-B1, EEEA-Net-C1, and EEEA-Net-N1 as small-scale architectures ($<$155 FLOPS). The large-scale architectures ($<$219 FLOPS) are EEEA-Net-A2, EEEA-Net-B2 EEEA-Net-C2, and EEEA-Net-N2.

\subsubsection{Architecture Evaluation on ImageNet dataset}

The discovery of architecture from Supernet is the separation of some layers from Supernet called subnets. Since Supernet and the subnets have different network architectures, the accuracy of the subnets, with pre-trained weight from Supernet, is very low when they were tested on the validation dataset. So, the subnets have calibrated the statistics of batch normalisation (BN) after searching on Supernet. The new BN statistics from the subnets were calculated using the validation dataset and updating the BN of all of the subnets. Thus, BN calibration can improve test accuracy value efficiency with the ImageNet dataset. 

Table~\ref{tab:5} shows the comparison of EEEA-Net performance with other models, using three main comparison factors: error rate, number of parameters, and \gls*{flops}. We classify models using architectural search methods such as auto, manual, or a combination. When comparing our small-scale architectures (EEEA-Net-A1, EEEA-Net-B1, EEEA-Net-C1, and EEEA-Net-N1) with GhostNet 1.0 \cite{han2020ghostnet}, we found that all our architectures outperform GhostNet 1.0. Also, EEEA-Net-A1, EEEA-Net-B1, EEEA-Net-C1, and EEEA-Net-N1 provide lower error and \gls*{flops} counts than MobileNetsV3 Large 0.75 \cite{howard2019searching}. However, the MobileNetsV3 Large 0.75 has fewer parameters than our models.

Similarly, when we compared our large-scale architectures with other architectures, we found that EEEA-Net-C2 ($\beta=7$) has a Top-1 error, and \gls*{flops} were lower than all other architectures, as shown in Table 4.  When we compare our architecture with MobileNetsV3 Large 1.0, EEEA-Net-C2 provides a 1\% less error value than MobileNetsV3 [28], and the \gls*{flops} count of EEEA-Net-C2 is reduced by 2 Million from MobileNetsV3. However, EEEA-Net-C2 had 0.6 Million more parameters than MobileNetsV3.

\begin{figure}%
    \centering
    \subfloat[\centering Top-1 accuracy vs FLOPS]{{\includegraphics[width=7cm]{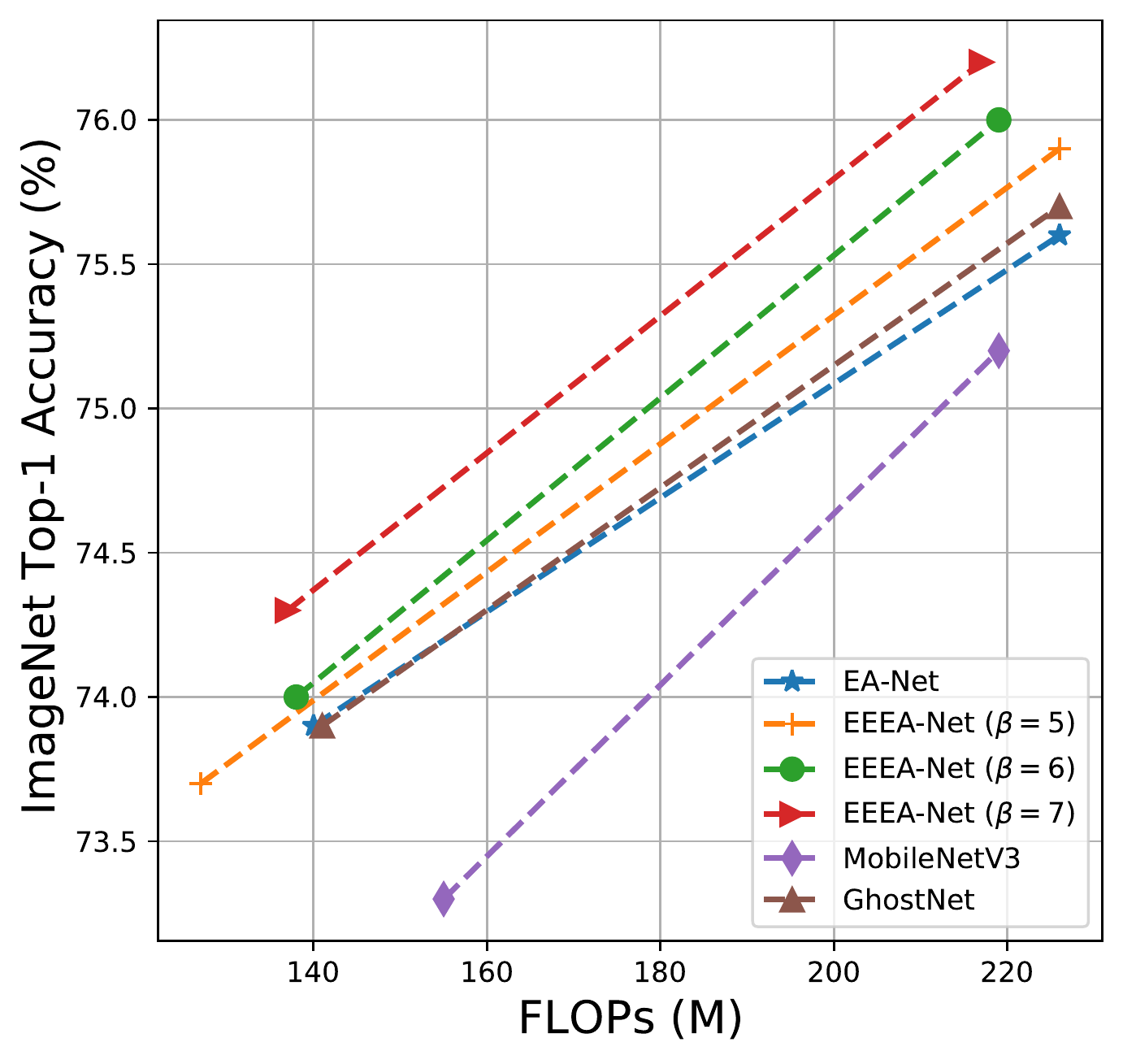} }}%
    \qquad
    \subfloat[\centering Top-1 accuracy vs Parameters]{{\includegraphics[width=7cm]{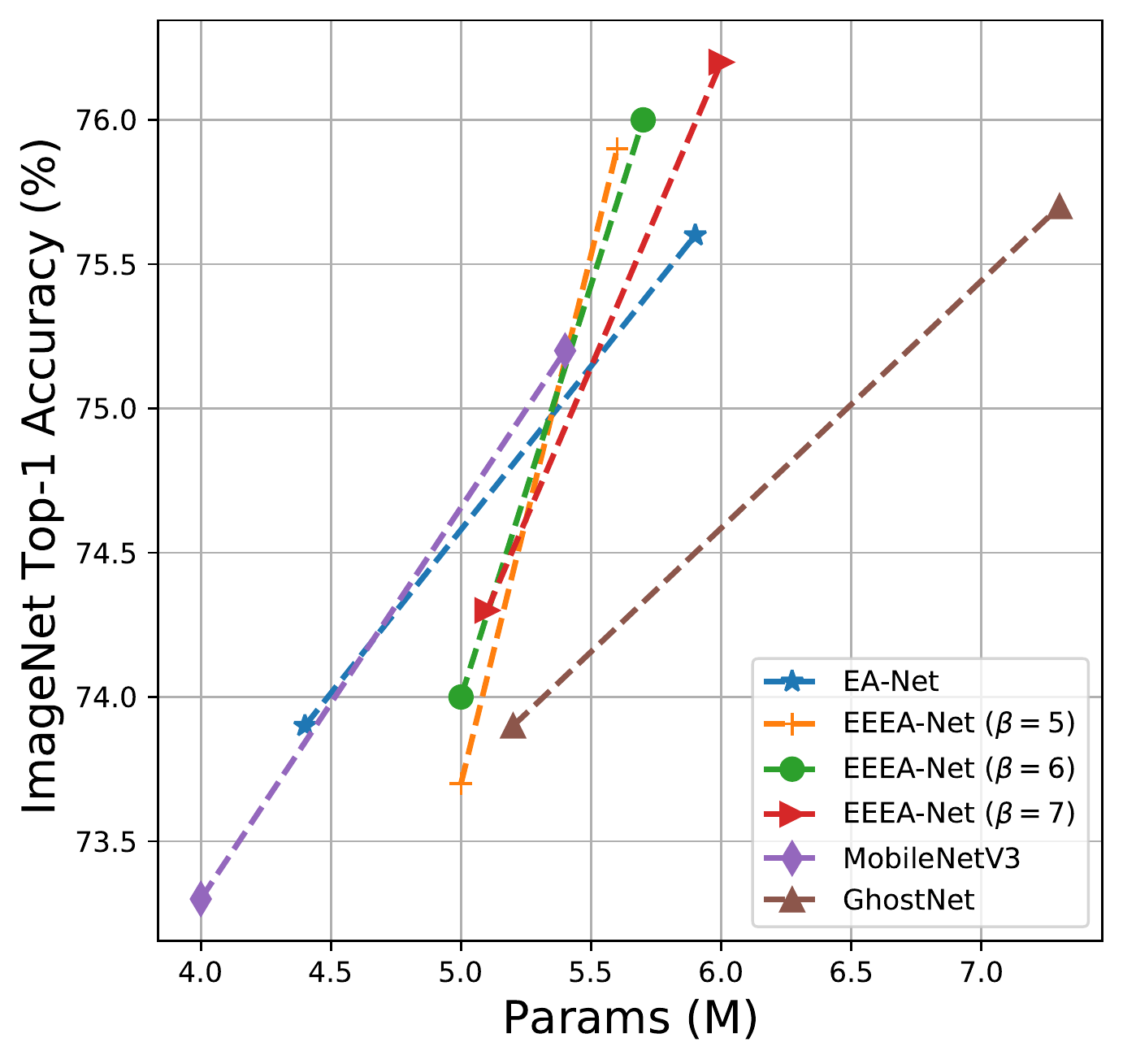} }}%
    \caption{Comparison of Top-1 accuracy, FLOPS (left), and parameters (right) between EEEA-Nets and MobileNetV3 [28] and GhostNet [40] on ImageNet dataset.}%
    \label{fig:8}%
\end{figure}

We chose MobileNetV3 and GhostNet, including both small and large versions, to compare with our architecture, as shown in Fig.~\ref{fig:8}. Overall, we observed that EEEA-Net-C ($\beta=7$) significantly outperforms MobileNetV3 and GhostNet for Top-1 accuracy and \gls*{flops}. Furthermore, EEEA-Net-C ($\beta=7$) achieves lower parameters than GhostNet. 

\subsection{Architecture Transfer}
After searching and evaluating the model using the ImageNet dataset for image recognition, the models trained with the ImageNet dataset can be further developed and applied to object detection, semantic segmentation and human keypoint detection applications.

\subsubsection{Object detection}

EEEA-Net-C2 ($\beta=7$) was used as the backbone for the object detection task to compare the effectiveness of our architecture on a real-world application. We utilised the same architecture trained with ImageNet datasets on the firmware called Single-Shot Detectors (SSD) \cite{liu2016ssd} and \gls*{yolov4} \cite{bochkovskiy2020yolov4}.

PASCAL VOC is a standard set of data used to measure an architecture's performance with object detection datasets. It consists of 20 classes, with bottles and plants being small objects with the lowest \gls*{ap} of all classes. We used the SSDLite framework \cite{sandler2018mobilenetv2} for fast and optimised processing on mobile devices. We also used the \gls*{yolov4} framework \cite{bochkovskiy2020yolov4} for high precision object detection.

All models were trained on the PASCAL VOC 2007 and VOC 2012 \cite{everingham2010the} train set by training the network for 200 epochs with a batch size of 32, SGD optimiser with weight decay equal to 0.0005 and momentum equal to 0.9, the initial learning rate is 0.01. It uses the scheduler with the cosine rule without a restart. All input images are resized to $320 \times 320$ pixels, and these models were used to evaluate the PASCAL VOC test set. 

For \gls*{yolov4}, we adopted MobileNet-V2 \cite{sandler2018mobilenetv2}, MobileNet-V3 \cite{howard2019searching} and EEEA-Net-C2 as the backbone of \gls*{yolov4}. All models were trained with 140 epochs and a batch size of 4. All inputs are random scales with multi-scale images ranging from 320 to 640 pixels. The label smoothing is 0.1, SGD optimiser with weight decay equal to 0.0005 and momentum equal to 0.9. The initial learning rate was set to 0.01 for a scheduler with a cosine rule with the warm-up strategy performed twice.


\begin{table}[t!]
\centering
\resizebox{\textwidth}{!}{%
\begin{tabular}{lcccccc}
\hline
\multicolumn{1}{c}{\multirow{2}{*}{Model}} &
  \multirow{2}{*}{Framework} &
  \multirow{2}{*}{Params   (M)} &
  \multirow{2}{*}{FLOPS  (M)} &
  \multicolumn{2}{c}{Small Objects AP (\%)} &
  \multirow{2}{*}{VOC2007 mAP (\%)} \\
\multicolumn{1}{c}{}   &         &                &               & Bottle        & Plant         &               \\ \hline
NASNet \cite{zoph2016neural}         & SSDLite & 5.22           & 1238          & 41.5          & 46.1          & 71.6          \\
DARTS \cite{liu2018darts}         & SSDLite & 4.73           & 1138          & 38.3          & 49.3          & 71.2          \\
ShuffleNet-V2 \cite{ma2018shufflenet} & SSDLite & \textbf{2.17}  & \textbf{355}  & 29.9          & 38.1          & 65.4          \\
MobileNet-V2 \cite{sandler2018mobilenetv2}  & SSDLite & 3.30           & 680           & 37.9          & 43.9          & 69.4          \\
MobileNet-V3 \cite{howard2019searching}  & SSDLite & 3.82           & 485           & 38.1          & 45.6          & 69.2          \\
MnasNet \cite{tan2019mnasnet}      & SSDLite & 4.18           & 708           & 37.7          & 44.4          & 69.6          \\
EEEA-Net-C2 (ours)     & SSDLite & 5.57           & 637           & \textbf{40.9} & \textbf{48.9} & \textbf{71.7} \\ \hline
MobileNet-V2 \cite{sandler2018mobilenetv2}  & YOLOv4  & 46.34          & 8740          & 66.4          & \textbf{58.4} & 81.5          \\
MobileNet-V3 \cite{howard2019searching}  & YOLOv4  & 47.30          & 8520          & 68.2          & 50.7          & 78.9          \\
EEEA-Net-C2 (ours)     & YOLOv4  & \textbf{31.15} & \textbf{5540} & \textbf{68.6} & 56.7          & \textbf{81.8} \\ \hline
\end{tabular}%
}
\caption{Result of Object detection with different backbones on PASCAL VOC 2007 test set.}
\label{tab:6}
\end{table}

Table~\ref{tab:6} shows the performance of our architecture for object detection. EEEA-Net-C2 achieved a higher \gls*{ap} than NASNet, DARTS, ShuffleNet-V2, MobileNet-V2, MobileNet-V3, and MnasNet for the SSDLite framework. Nonetheless, EEEA-Net-C2 has 152 more million \gls*{flops} than MobileNet-V3. For fairness, we used MobileNet-V2, MobileNet-V3 and EEEA-Net-C2 for training and evaluated these models using the PASCAL VOC test dataset via the \gls*{yolov4} framework. The EEEA-Net-C2 significantly outperformed both MobileNet-V2 and MobileNet-V3.

\subsubsection{Semantic Segmentation}

The cityscape dataset \cite{cordts2016the} was chosen to experiment with semantic segmentation. It is a large-scale dataset of street scenes in 50 cities. Cityscapes provide dense pixel annotations of 5,000 images. These images were divided into three groups of 2,975, 500, 1,525 images for training, validation, and testing. We used BiSeNet \cite{yu2018bisenet} with different backbones to evaluate our architecture's performance for semantic segmentation on the Cityscapes dataset. NASNet, DARTS, ShuffleNet-V2, MobileNet-V2, MobileNet-V3, MnasNet, and EEEA-Net-C2 have trained 80,000 iterations with a poly learning scheduler at an initial learning rate of 0.01 and batch size equals 16. All training images were resized to $1024 \times 1024$ pixels using image augmentation using colour jitter, random scale, and random horizontal flip.

Table~\ref{tab:7} shows that ShuffleNet-V2 achieved a smaller number of parameters and lower \gls*{flops} than other architectures. However, MobileNet-V2 achieved a greater \gls*{miou} than ShuffleNet-V2, MobileNet-V3, MnasNet, and EEEA-Net-C2. The \gls*{miou} of EEEA-Net-C2 is the same as MnasNet. It is better than ShuffleNet-V2 and MobileNet-V3. 


\begin{table}[t!]
\centering
\begin{tabular}{lccc}
\hline
\multicolumn{1}{c}{Model} & Params (M)    & FLOPS (G)      & mIoU (\%)     \\ \hline
NASNet \cite{zoph2016neural}            & 7.46          & 36.51          & 77.9          \\
DARTS \cite{liu2018darts}          & 6.64          & 34.77          & 77.5          \\ \hline
ShuffleNet-V2 \cite{ma2018shufflenet}    & \textbf{4.10} & \textbf{26.30} & 73.0          \\
MobileNet-V2 \cite{sandler2018mobilenetv2}     & 5.24          & 29.21          & \textbf{77.1} \\
MobileNet-V3 \cite{howard2019searching}    & 5.60          & 27.09          & 75.9          \\
MnasNet \cite{tan2019mnasnet}         & 6.12          & 29.50          & 76.8          \\
EEEA-Net-C2 (ours)        & 7.34          & 28.65          & 76.8          \\ \hline
\end{tabular}
\caption{Results of BiSeNet with different backbones on Cityscapes validation set. (single scale and no flipping).}
\label{tab:7}
\end{table}

\subsubsection{Keypoint Detection}

Human keypoint detection, also known as human pose estimation, is the visual sensing of human gestures from keypoints such as the head, hips, or ankles. MS COCO \cite{lin2014microsoft} is a comprehensive dataset to measure keypoint detection performance, consisting of data for 250,000 persons, with the data labelle at 17 keypoints. SimpleBaseline \cite{xiao2018simple} is a framework for keypoint detection, enabling easier changes to backbones. Given this, it allowed us to adapt to other architectures more simply.

All architectures were trained on the MS COCO train2017 set by training the network for 140 epochs with a batch size of 128, Adam optimiser, the initial learning rate is 0.001, which is reduced to 0.0001 at 90th epoch and reduced to 0.00001 at 120th epoch. The training set is resized to $256 \times 192$ pixels using random rotation, scale, and flipping data.

Table~\ref{tab:8} shows the experimental result of SimpleBaseline with different backbones. Our EEEA-Net-C2 performed better than other backbones in the number of parameters. As well, EEEA-Net-C2 outperforms small architectures (excluding NASNet and DARTS).


\begin{table}[t!]
\centering
\begin{tabular}{lccc}
\hline
\multicolumn{1}{c}{Model} & Params (M)    & FLOPS (M)       & AP (\%)       \\ \hline
NASNet \cite{zoph2016neural}            & 10.66         & 569.11          & 67.9          \\
DARTS \cite{liu2018darts}            & 9.20           & 531.77          & 66.9          \\ \hline
ShuffleNet-V2 \cite{ma2018shufflenet}    & 7.55          & \textbf{154.37} & 60.4          \\
MobileNet-V2 \cite{sandler2018mobilenetv2}     & 9.57          & 306.80           & 64.9          \\
MobileNet-V3 \cite{howard2019searching}     & 9.01          & 223.16          & 65.3          \\
MnasNet \cite{tan2019mnasnet}          & 10.45         & 320.17          & 62.5          \\
EEEA-Net-C2 (ours)        & \textbf{7.47} & 297.49          & \textbf{66.7} \\ \hline
\end{tabular}
\caption{Results of SimpleBaseline with different backbone settings on MS COCO2017 validation set. Flip is used during validation.}
\label{tab:8}
\end{table}

\subsection{Limitations}

The development of a \gls*{nas} search with only one GPU processor was a challenge for the reasons set out below. Setting the appropriate number of populations, the number of generations of each population, and the number of search epochs suitable for one GPU processor presents considerable difficulties. All of these parameters affect the model search time. 

Increasing the number of generations increases the computing cost, but increasing the number of generations provides an opportunity for greater recombination of populations, thereby maximising the efficiency of discovering new populations. Moreover, an increased number of search epoch helps improve each population's error fitness value. 

All these restrictions help to improve the \gls*{nas} search. However, increasing these numbers influences search time.  For example, increasing the number of search epochs from 1 epoch to 10 epochs results in a 10$\times$ increase in search time. 

\section{Conclusion}

We achieved our research goals by successfully developing a \gls*{cnn} architecture suitable for an on-device processor with limited computing resources and applying it in real-world applications. 

This outcome was achieved by significantly reducing the computational cost of a neural architecture search. We introduced the Early Exit Population Initialisation (EE-PI) for Evolutionary Algorithm method to create the EEEA-Nets model. Our method achieved a massive reduction in search time on CIFAR-10 dataset; 0.34 to 0.52 GPU days. This must be seen as an outstanding outcome compared against other state-of-the-art models, such as the NSGA-Net model, which required 4 GPU days, the 2,000 GPU days of the NASNet model and the 3,150 GPU days of the AmoebaNet model.

In the EEEA-Nets architecture, our emphasis was on reducing the number of parameters, the error rate and the computing cost. We were able to achieve this by introducing an Early Exit step into the Evolutionary Algorithm. 

Our EEEA-Nets architectures were searched on image recognition task, transferring architectures to other tasks. Experimentally, EEEA-Net-C2 is significantly better than MobileNet-V3 on image recognition, object detection, semantic segmentation, and keypoint detection tasks. Addressing this latter task had not been achieved or even attempted in any other \gls*{cnn} model. Therefore, our architectures can be deployed on devices with limited memory and processing capacity by achieving these significant reductions, allowing real-time processing on smartphones or on-device systems. 

The task of optimising the search for multi-objective evolutionary algorithms shall be continued as our future work to find better-performing models. In addition, we will consider applying a multi-objective evolutionary algorithm with \gls*{eepi} to find mobile-suitable models in other applications such as marine detection or pest detection.

\section*{Acknowledgements}

The authors would like to acknowledge the Thailand Research Fund's financial support through the Royal Golden Jubilee PhD. Program (Grant No. PHD/0101/2559). The study was undertaken using the National Computational Infrastructure (NCI) in Australia under the National Computational Merit Allocation Scheme (NCMAS). Further, we would like to extend our appreciation to Mr Roy I. Morien of the Naresuan University Graduate School for his assistance in editing the English grammar, syntax, and expression in the paper.

\bibliography{mybibfile}

\section{Appendix}

\subsection{Architecture Visualisation}

\begin{figure}[b!]
\centering

\subfloat[EA-Net ($\beta=0$) Normal Cell.]{\includegraphics[width=0.45\linewidth]{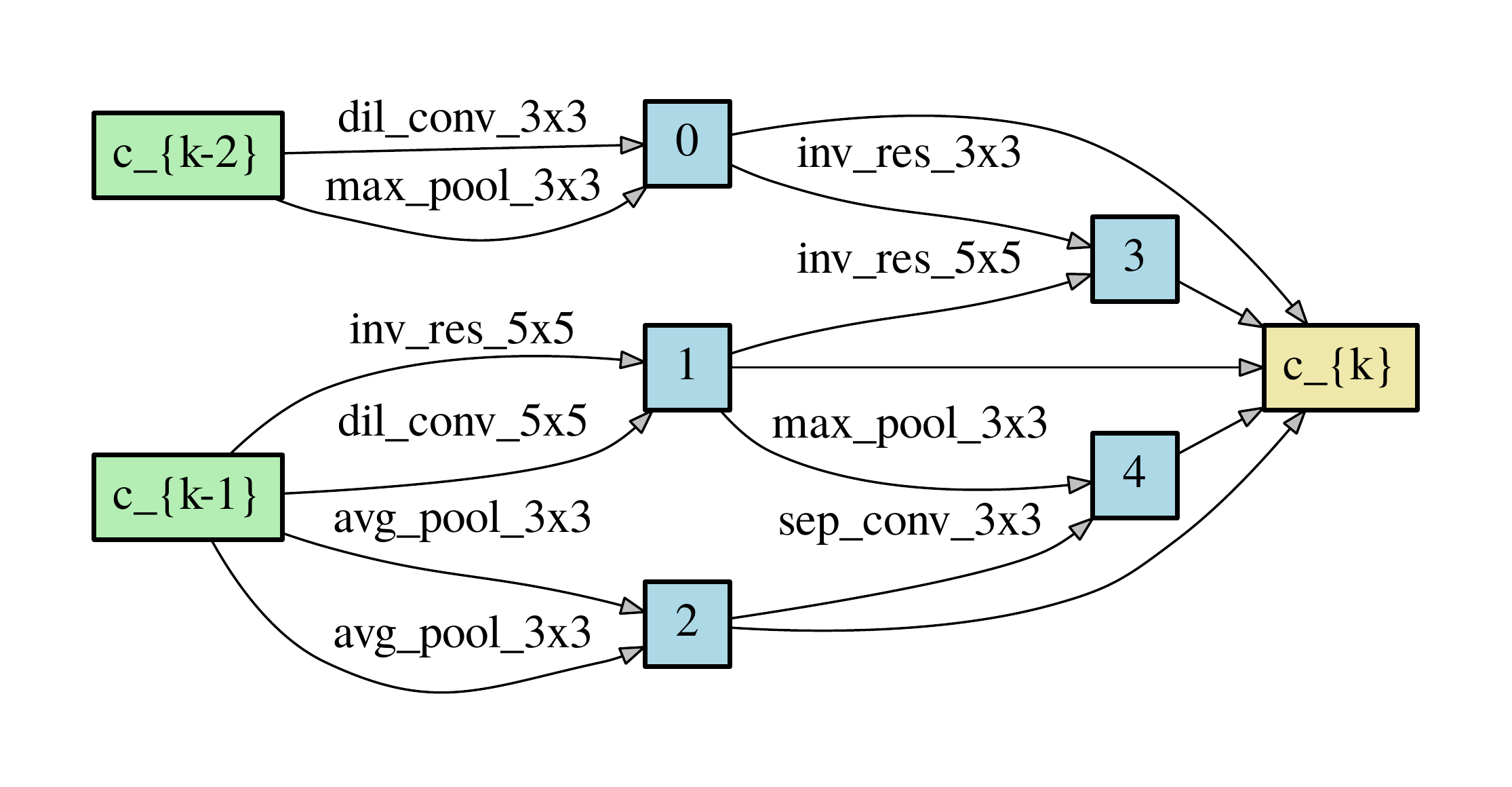}
\label{fig:cross_a}}
\qquad
\subfloat[EA-Net ($\beta=0$) Reduction Cell.]{\includegraphics[width=0.45\linewidth]{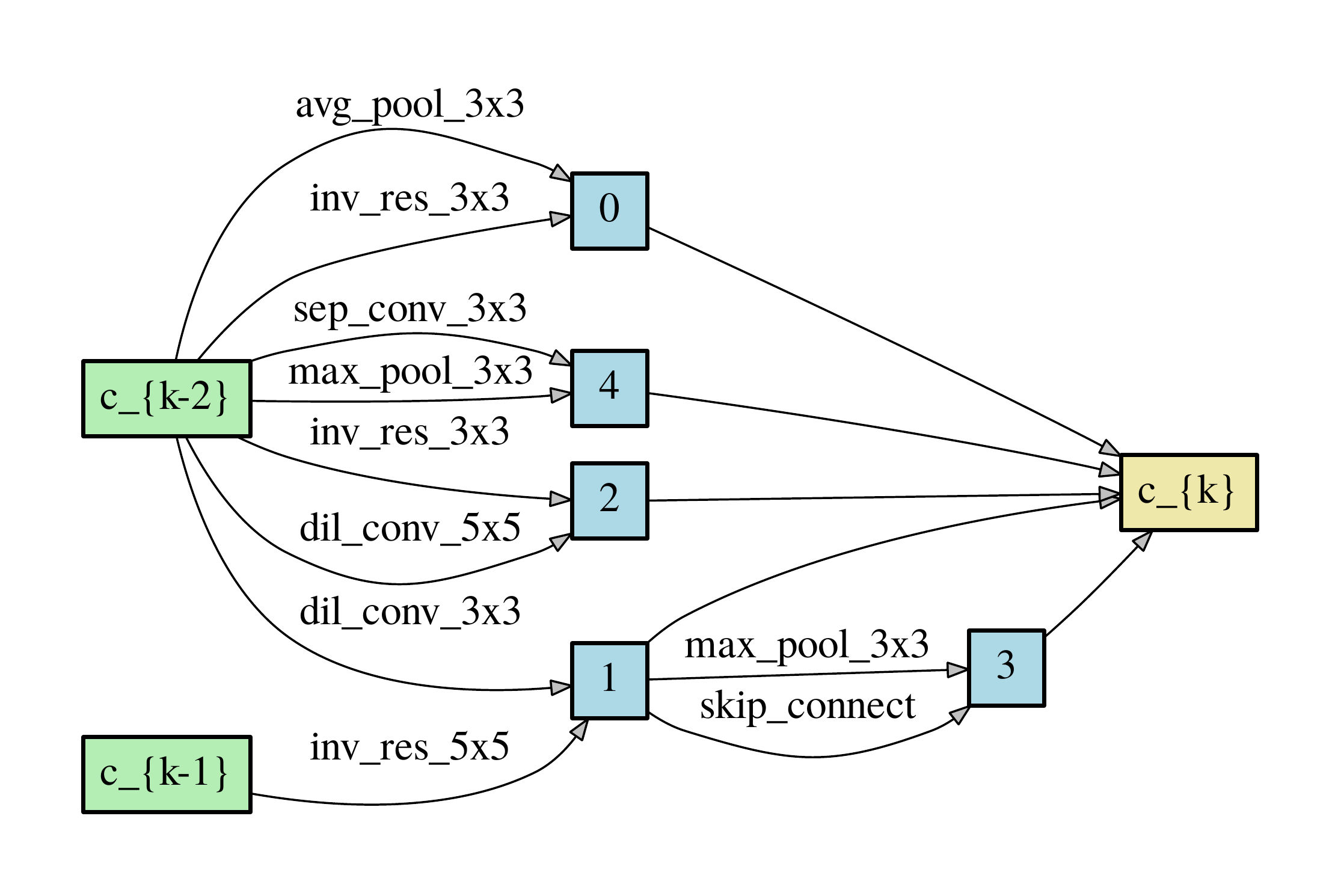}
\label{fig:cross_b}}
\qquad

\subfloat[EEEA-Net-A ($\beta$=3.0) Normal Cell.]{\includegraphics[width=0.45\linewidth]{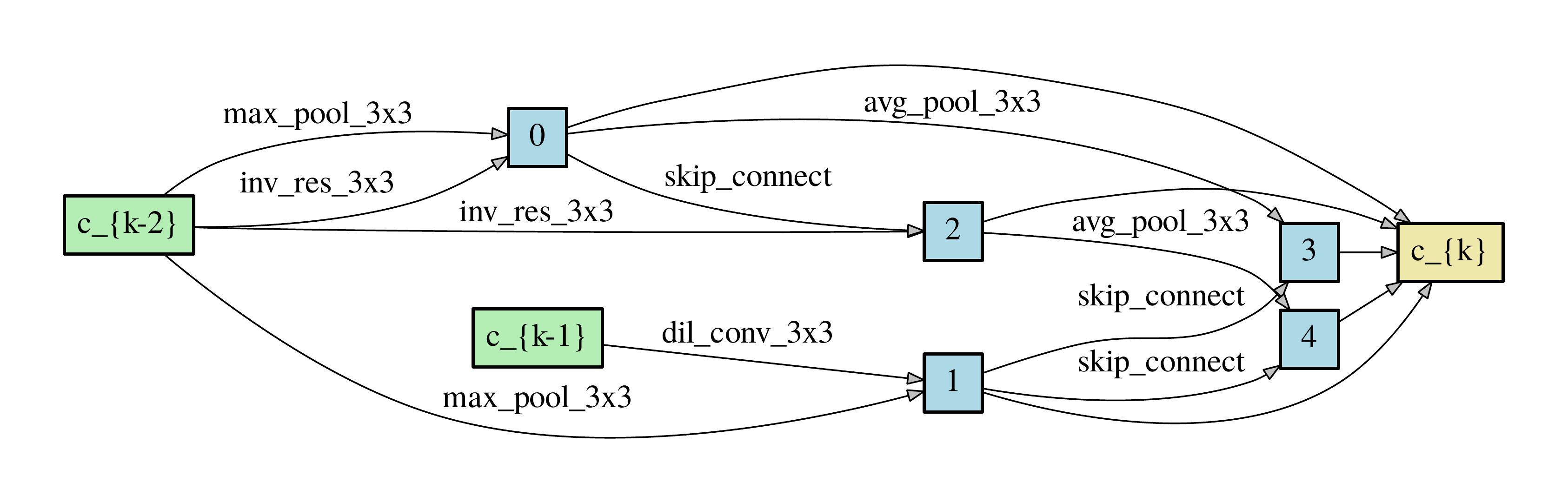}
\label{fig:cross_a}}
\qquad
\subfloat[EEEA-Net-A ($\beta$=3.0) Reduction Cell.]{\includegraphics[width=0.45\linewidth]{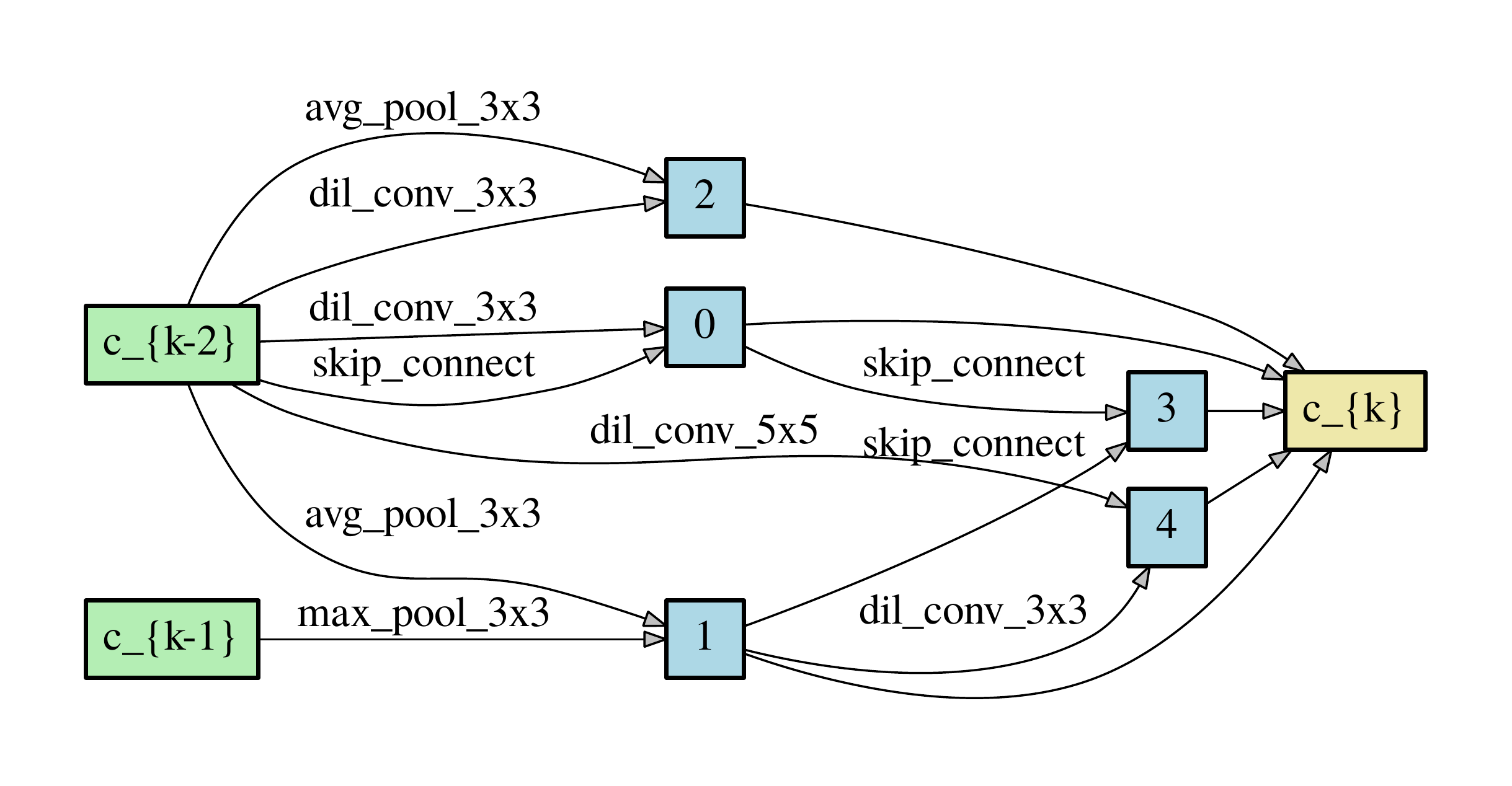}
\label{fig:cross_b}}
\qquad

\subfloat[EEEA-Net-B ($\beta$=4.0) Normal Cell.]{\includegraphics[width=0.45\linewidth]{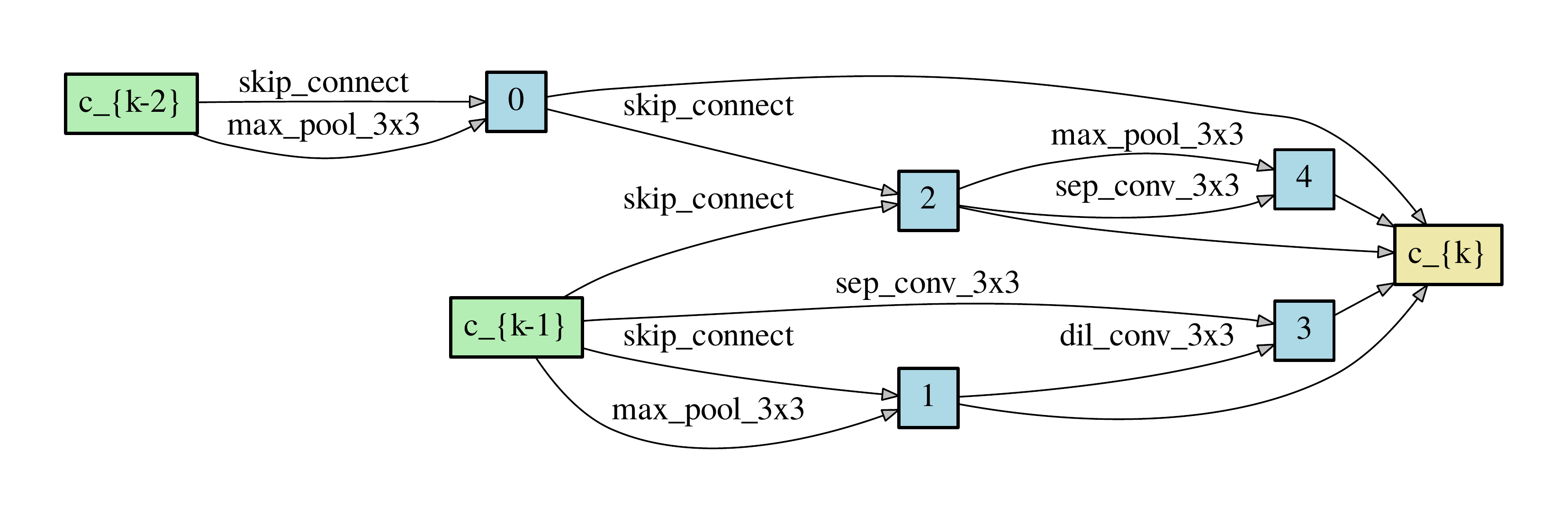}
\label{fig:cross_a}}
\qquad
\subfloat[EEEA-Net-B ($\beta$=4.0) Reduction Cell.]{\includegraphics[width=0.45\linewidth]{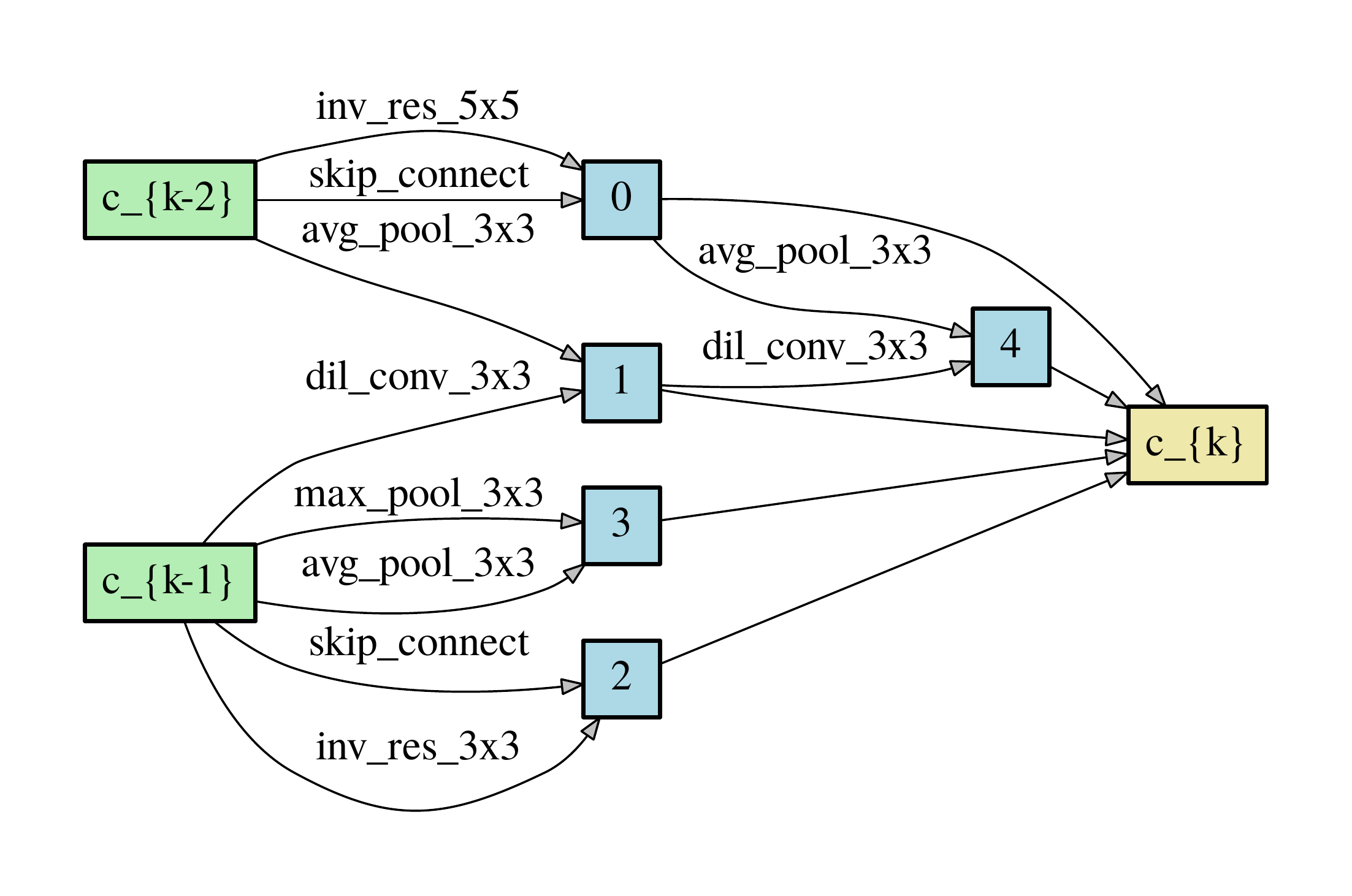}
\label{fig:cross_b}}
\qquad

\subfloat[EEEA-Net-C ($\beta$=5.0) Normal Cell.]{\includegraphics[width=0.45\linewidth]{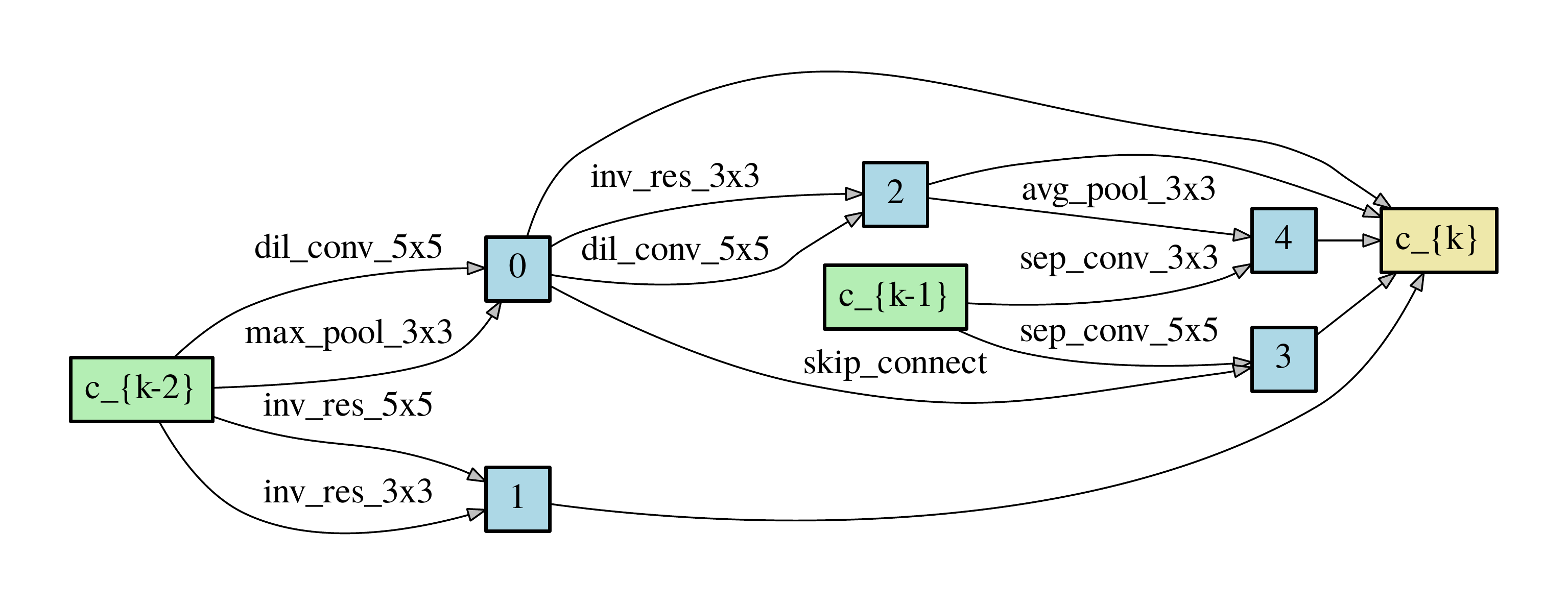}
\label{fig:cross_a}}
\qquad
\subfloat[EEEA-Net-C ($\beta$=5.0) Reduction Cell.]{\includegraphics[width=0.45\linewidth]{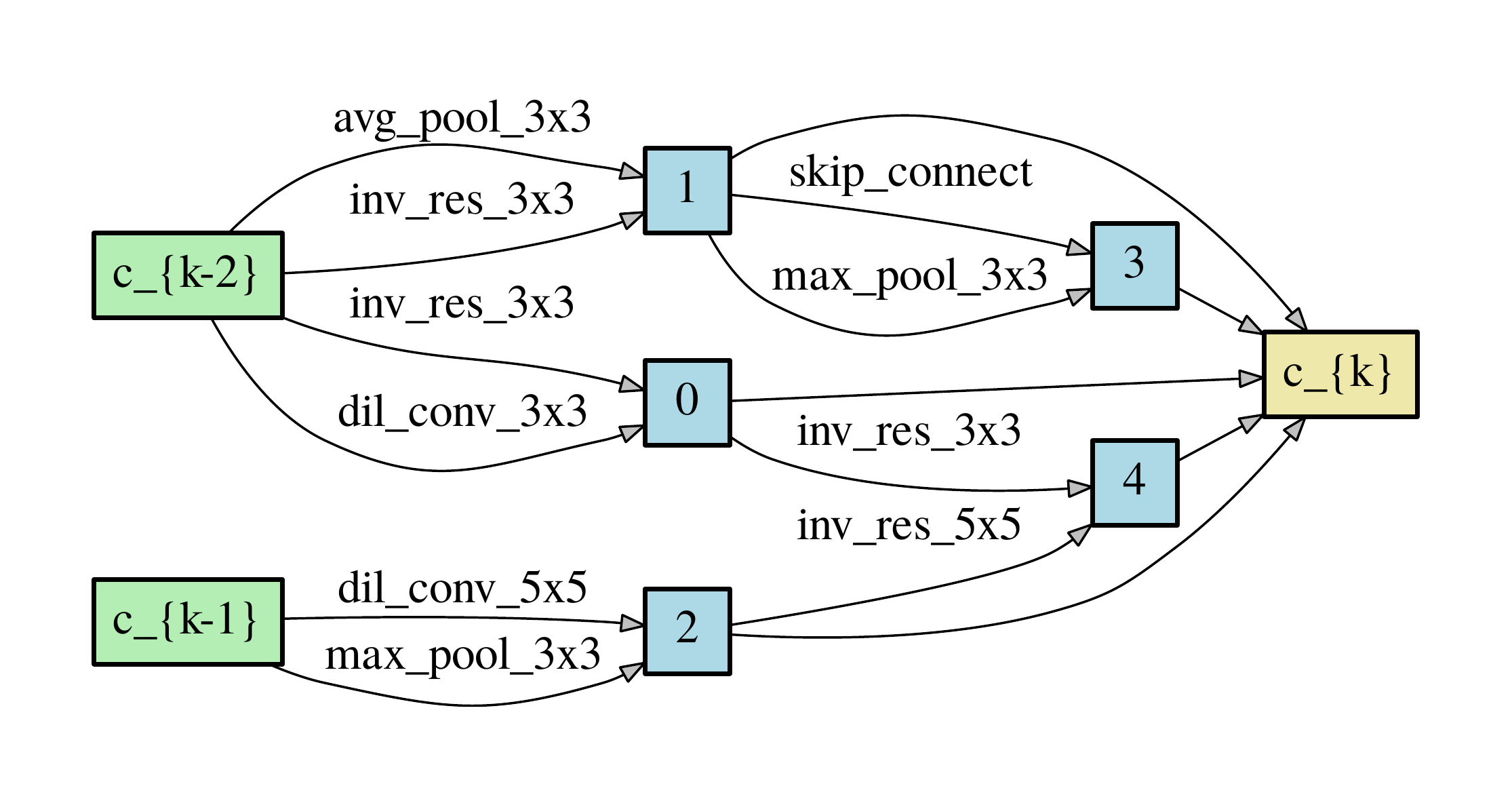}
\label{fig:cross_b}}
\qquad

\caption{Normal and Reduction cells learned on CIFAR-10: EA-Net ($\beta$=0.0), EEEA-Net-A ($\beta$=3.0), EEEA-Net-B ($\beta$=4.0), and EEEA-Net-C ($\beta$=5.0).}
\label{fig:9}
\end{figure}

\begin{figure}[]
\centering\includegraphics[width=0.90\linewidth]{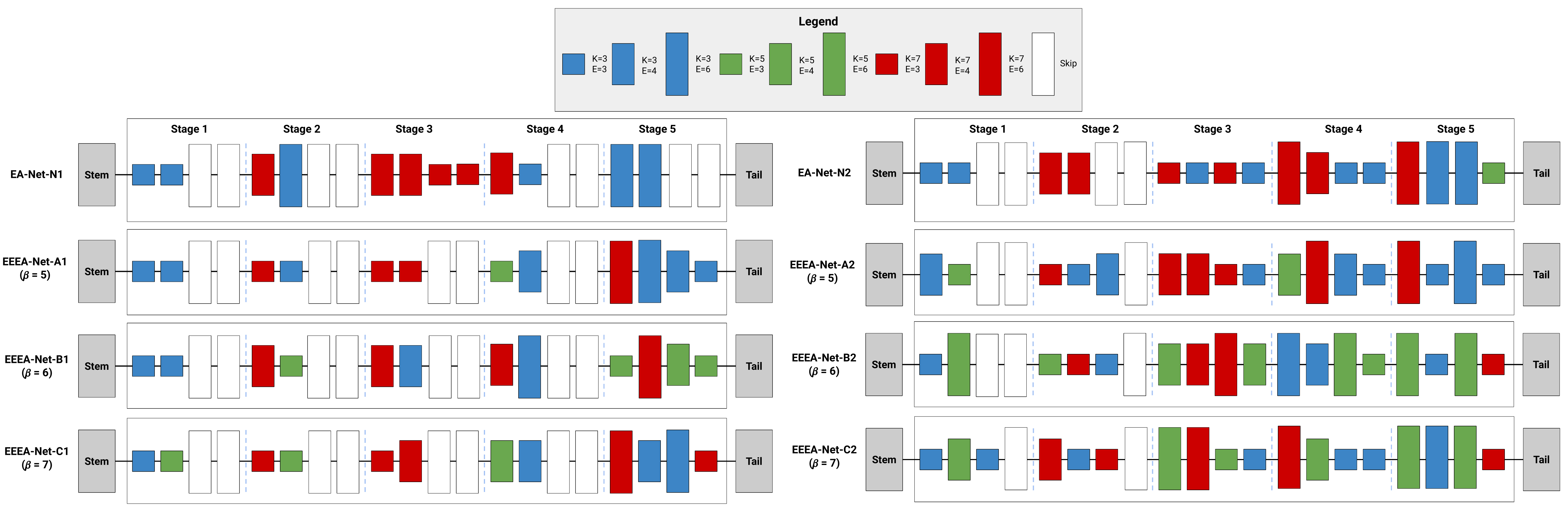}
\caption{EA-Nets and EEEA-Nets architectures were searched from ImageNet datasets. The stem and tail layers in all architectures are the same.}
\label{fig:10}
\end{figure}

This section visualises the architectures obtained by searching for EEEA-Nets with CIFAR-10 datasets, as shown in Fig.~\ref{fig:9}, and EEEA-Nets with ImageNet datasets shown in Fig.~\ref{fig:10}. These architectures are the most reliable, minimising three goals.


\subsection{Error Analysis of EEEA-Net-C2}
Our experiment applied EEEA-Net-C2 for detection, semantic segmentation, and human keypoint detection, where we concluded that the EEEA-Net-C2 model was better than the MobileNet-V3 model. For error analysis of the EEEA-Net-C2 model, images for each application were processed to check the correct results. In this appendix, error analysis of the EEEA-Net-C2 model is divided into three parts: error analysis of object detection, error analysis of semantic segmentation errors and error analysis of human keypoint detection.

\subsubsection{Object detection}

\begin{figure}[]
\centering\includegraphics[width=0.90\linewidth]{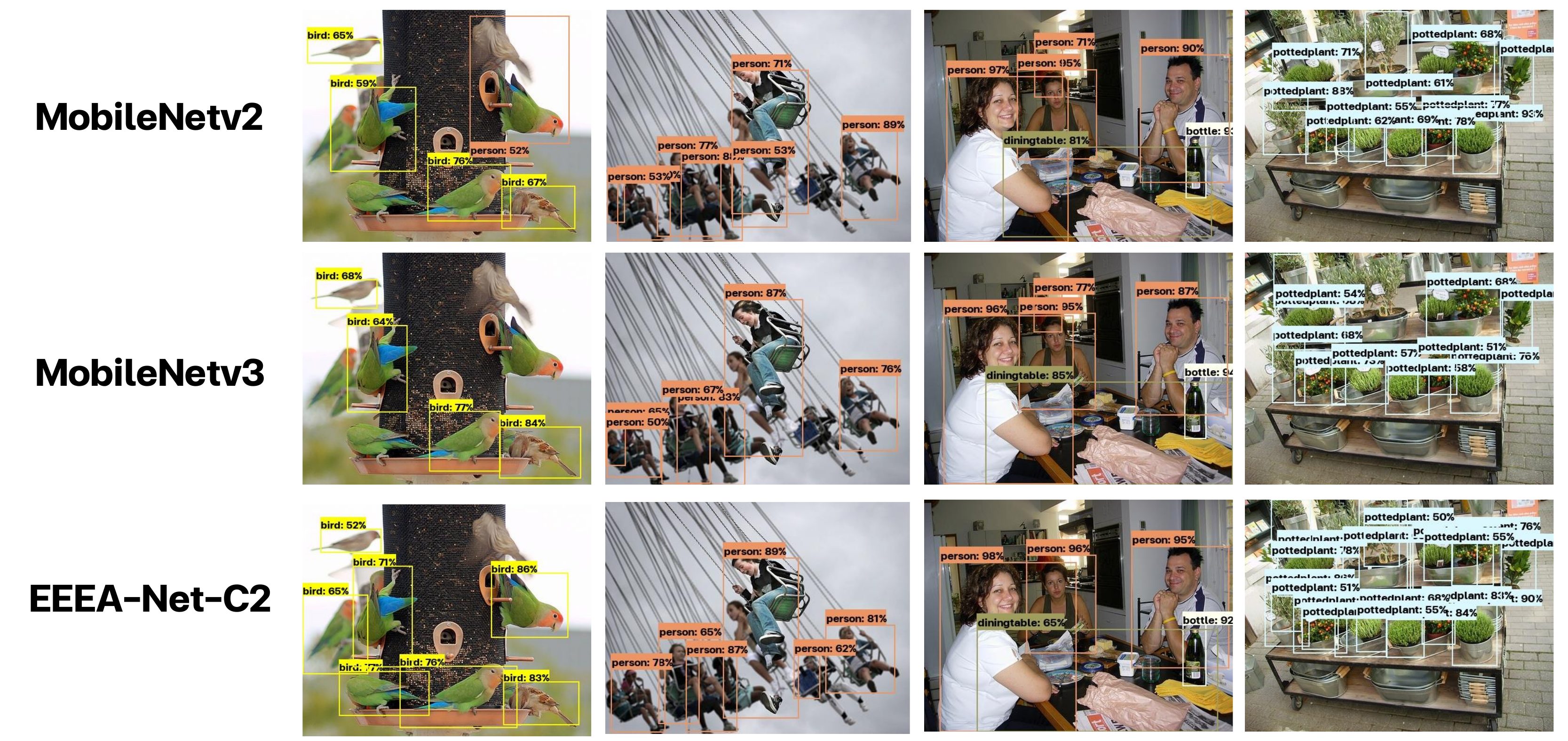}
\caption{An example of object detection results of MobileNetv2, MobileNetv3, and EEEA-Net-C2 models.}
\label{fig:11}
\end{figure}

Object detection results from MobileNetv2, MobileNetv3 and EEEA-Net-C2 models are shown in Fig.~\ref{fig:11}. Given the error from the images in the first column, the MobileNetv2 model was found to have mistakenly identified ``bird" as ``person", while the MobileNetv3 model was unable to find ``bird". However, the EEEA-Net-C2 model detected the location of all ``birds". 

As with the second column in Fig.~\ref{fig:11}, the EEEA-Net-C2 model can identify all ``persons" positions. However, only the EEEA-Net-C2 model could not locate the hidden ``person" behind the middle woman in the third column image. Additionally, in the fourth column pictures, the EEEA-Net-C2 model was able to identify more ``plant pots" than the MobileNetv2 and MobileNetv3 models.

\subsubsection{Semantic segmentation}

\begin{figure}[]
\centering\includegraphics[width=0.90\linewidth]{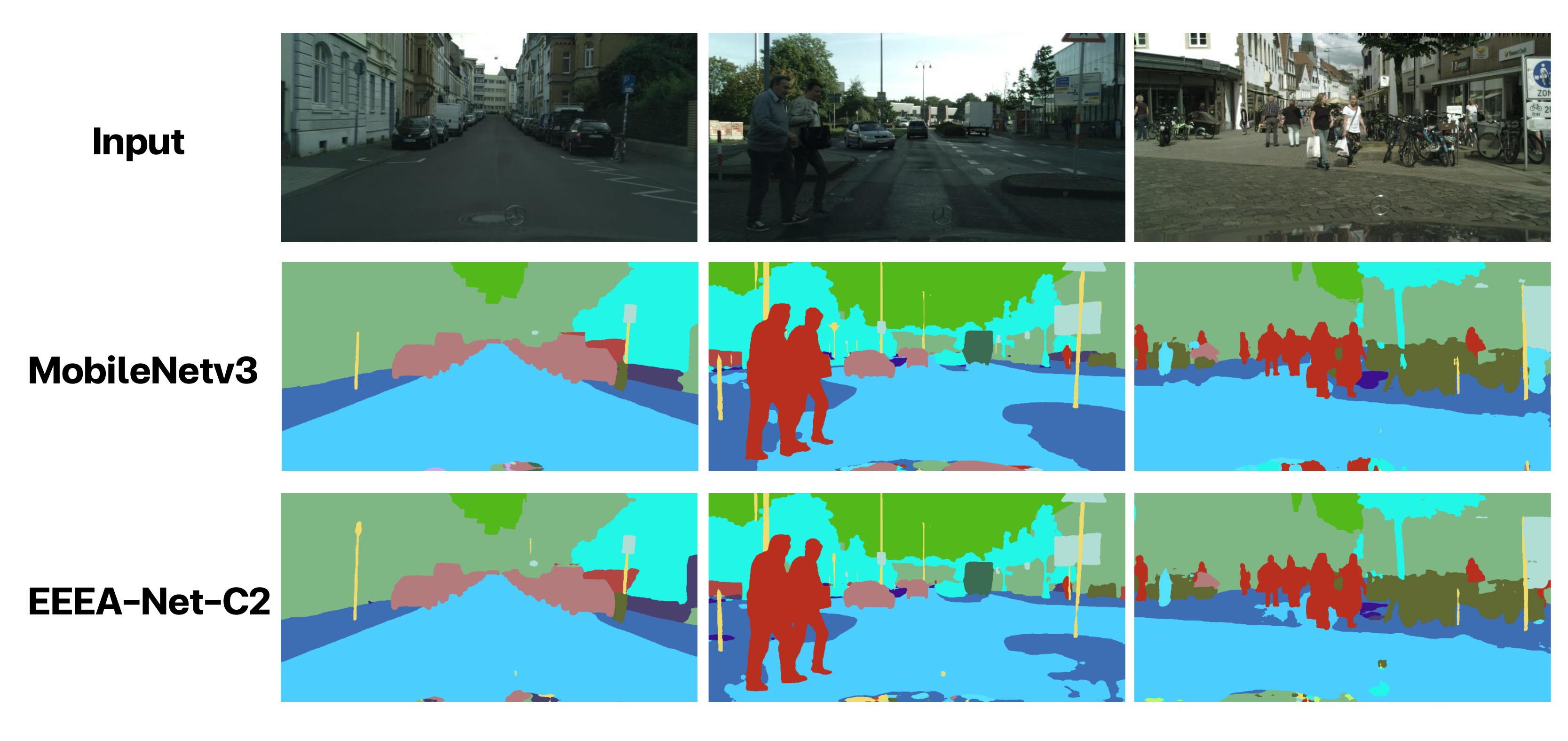}
\caption{An example of semantic segmentation results of MobileNetv2, MobileNetv3 and EEEA-Net-C2 models.}
\label{fig:12}
\end{figure}

The results of visual image segmentation using MobileNetv3 and EEEA-Net-C2 models are shown in Fig.~\ref{fig:12}.  The error of the semantic segmentation results can be determined from the pictures of the first column. It was found that MobileNetv3 models could segment only ``Traffic sign pole". However, the MobileNetv3 model cannot segment the left ``traffic sign", while the EEEA-Net-C2 model can segment both ``pole and sign". 

The pictures in the second column from Fig.~\ref{fig:12} depicts that the EEEA-Net-C2 model segmented from the ``traffic island" was less than the MobileNetv3 model. Next, in the third column, the EEEA-Net-C2 model segmented the ``footpath" more precisely than the MobileNetv3 model.

\subsubsection{Human keypoint detection}

\begin{figure}[]
\centering\includegraphics[width=0.90\linewidth]{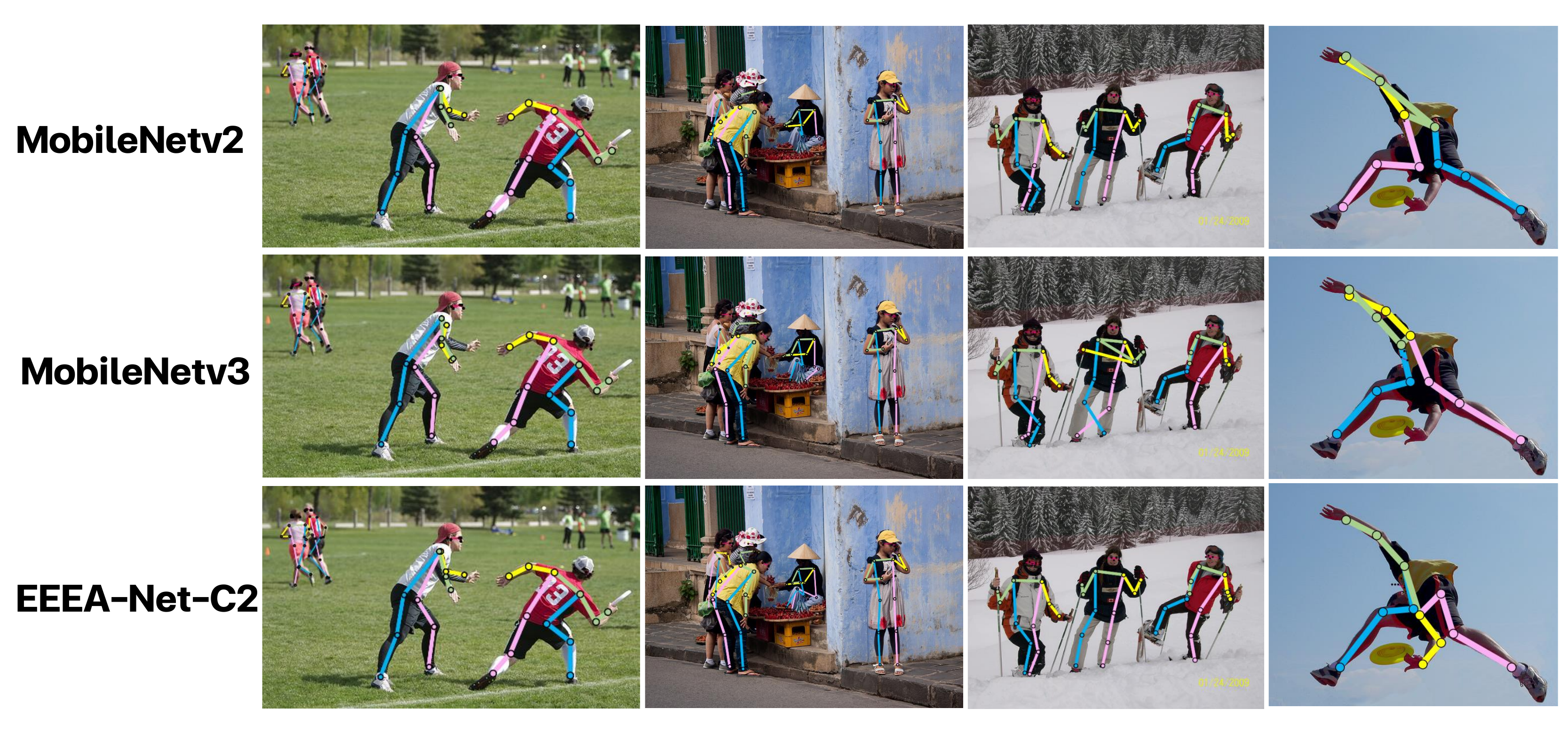}
\caption{An example of human keypoint detection results of MobileNetv2, MobileNetv3 and EEEA-Net-C2 models.}
\label{fig:13}
\end{figure}

The human keypoint detection results from the MobileNetv2, MobileNetv3 and EEEA-Net-C2 models are shown in Fig.~\ref{fig:13}. When considering the error from the pictures in the first column, the MobileNetv3 model was found to indicate the ``left arm" position, while the MobileNetv2 and EEEA-Net-C2 models were able to locate.

Only the MobileNetv3 model can pinpoint the ``leg" of the person sitting in the second column pictures. However, in the third column pictures, the EEEA-Net-C2 model can locate the middle person's ``arms and legs", while the MobileNetv3 model identifies the person's wrong location. Additionally, in the fourth column pictures, the EEEA-Net-C2 model could locate the ``arm and leg" more accurately than the MobileNetv2 and MobileNetv3 models.

The above data shows that the EEEA-Net-C2 model is accurate as well as inaccurate. The EEEA-Net-C2 model was designed and searched with the ImageNet dataset. Thus, the EEEA-Net-C2 model may have errors when used with other dataset or tasks. However, the EEEA-Net-C2 model has a performance higher than the MobileNetv2 and MobileNetv3 models on the same dataset and framework used in the three applications.

\subsection{Mobile Processing}
This appendix measures the performance of our EEEA-Net-C2 model and other state-of-the-art models on the smartphone and only CPU. All trained models with the ImageNet dataset are converted to the PyTorch JIT version to enable easy implementation on different platforms.

\begin{figure}[]
\centering\includegraphics[width=0.90\linewidth]{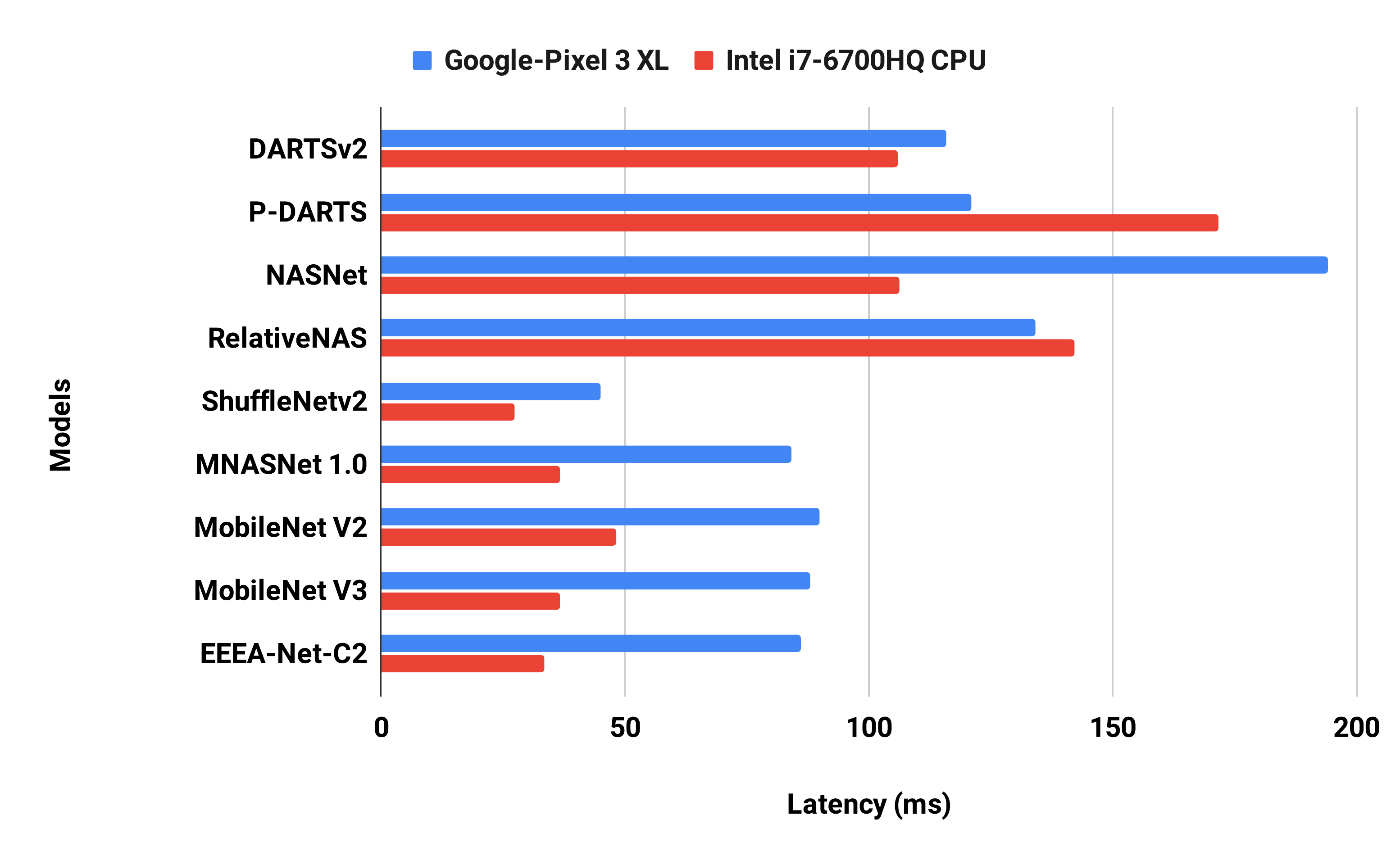}
\caption{Comparison of latency between EEEA-Net and other state-of-the-art models on non-GPU processing.}
\label{fig:14}
\end{figure}

Fig.~\ref{fig:14} shows the latency performance with 100 images with 224x224 pixels on the Google Pixel 3 XL smartphone (blue bars) and Intel i7-6700HQ CPU (red bars) devices with non-GPU resources by DARTSv2, P-DARTS, NASNet, ReletiveNAS, ShuffleNetV2, MNASNet 1.0, MobileNetV2, MobileNetV3, and EEEA-Net-C2 models. 

On the Google Pixel 3 XL, the EEEA-Net-C2 model processed each image in 86 milliseconds per image, whereas the MobileNetV2 and MobileNetV3 models took 90 and 88 milliseconds, respectively. The EEEA-Net-C2 model has a shorter latency time than state-of-the-art models (including DARTSv2, P-DARTS, NASNet, and ReletiveNAS), and MobileNets models, primarily models for smartphones, are compared to EEEA-Net-C2. 

Also, on the Intel i7-6700HQ CPU, the latency time of the EEEA-Net-C2 model has shorter latency than state-of-the-art models and lightweight models (including MNASNet 1.0, MobileNetV2, and MobileNetV3). 

\subsection{NAS-Bench dataset}

Experimental results with CIFAR-10, CIFAR-100, and ImageNet datasets were compared between NAS methods. The results obtained from different methods are achieved with different settings such as hyperparameters (e.g., learning rate and batch size), data augmentation (e.g., Cutout and AutoAugment). Thus, the comparison may not be fair.

This section has implemented an Early Exit method to model search from the NAS-Bench datasets, which avoids unfair comparisons and provides a uniform benchmark for NAS algorithms. The dataset used in this experiment was NAS-Bench-101, NAS-Bench-1Shot1 and NAS-Bench-201.

\subsubsection{NAS-Bench-101}

The NAS-Bench-101 \cite{ying2019nas} provides a table dataset of 423,624 unique architectures. These architectures have trained and evaluated the CIFAR-10 dataset to allow our work to search and query the mapped performance in the dataset in a few milliseconds. 

\begin{figure}[t!]
\centering\includegraphics[width=0.75\linewidth]{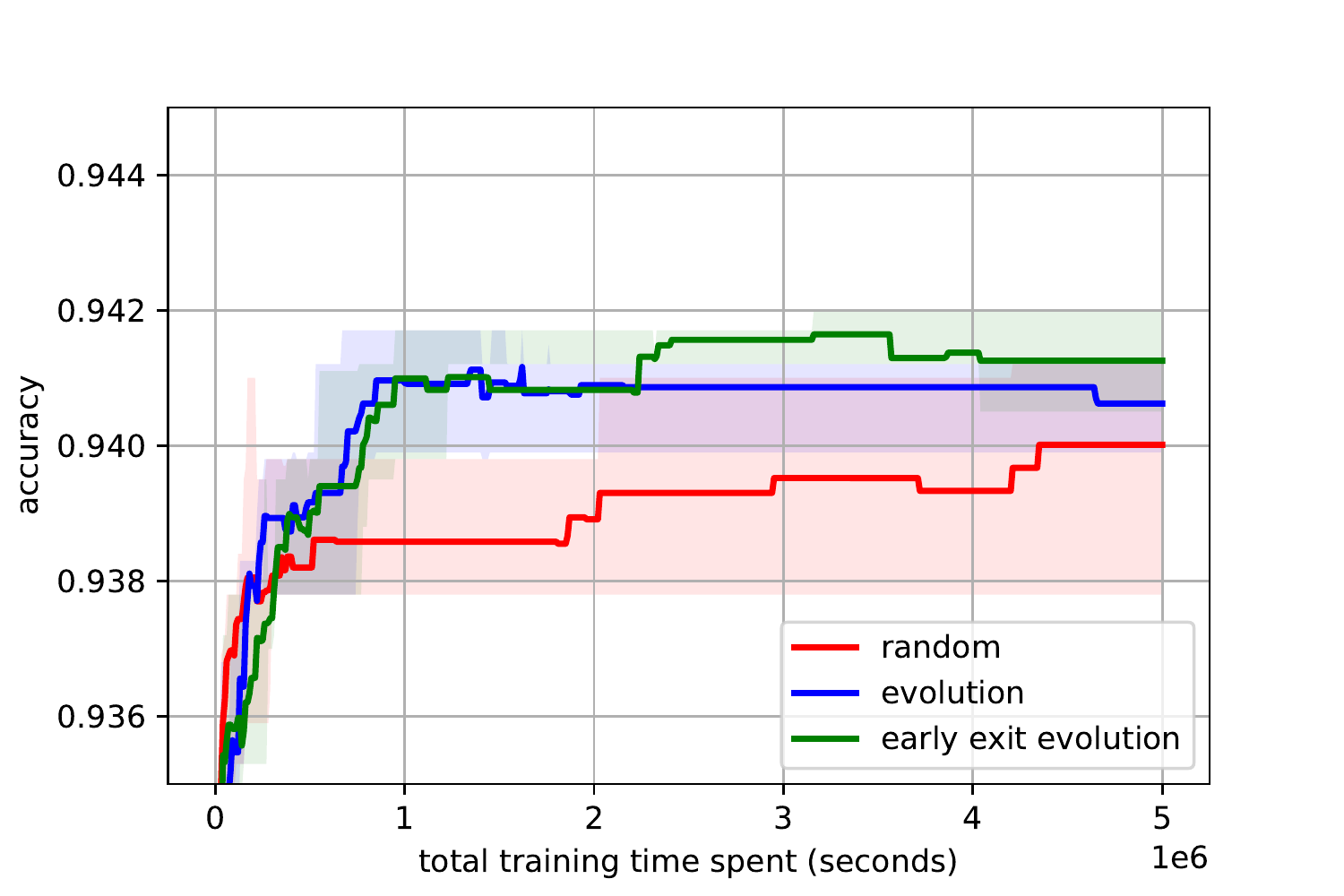}
\caption{Comparison of accuracy between random search, regularised evolution and Early Exit evolution algorithms on NAS-Bench-101.}
\label{fig:15}
\end{figure}

We have re-implemented model search from the NAS-Bench-101 dataset by using random search, regularised evolution and Early Exit evolution algorithms to search and query the performance of the resulting models. We used a re-implemented regularised evolution with the Early Exit method by taking population size of 100, a tournament size of 10, and maximum parameters' Early Exit ($\beta$) is 25 million.

The results in Fig.~\ref{fig:15} show that our early exit evolution algorithm tends to be higher in accuracy than the regularised evolution from 2 million seconds to 5 million seconds. Overall, the regularised evolution algorithm appears to perform better than the random search algorithm. However, our early exit evolution tends to outperform both random search and regularised evolution algorithms.

\subsubsection{NAS-Bench-1Shot1}
NAS-Bench-1Shot1 \cite{zela2020nas} is the benchmark for a one-shot neural architecture search, developed from the NAS-Bench-101 search space by tracking the trajectory and performance of the obtained architectures for three search spaces: 6,240 architectures for search space 1, 29,160 architectures for search space 2, and 363,648 architectures for search space 3.

\begin{figure}[t!]
   \subfloat[\label{genworkflow}]{%
      \includegraphics[width=0.3\textwidth]{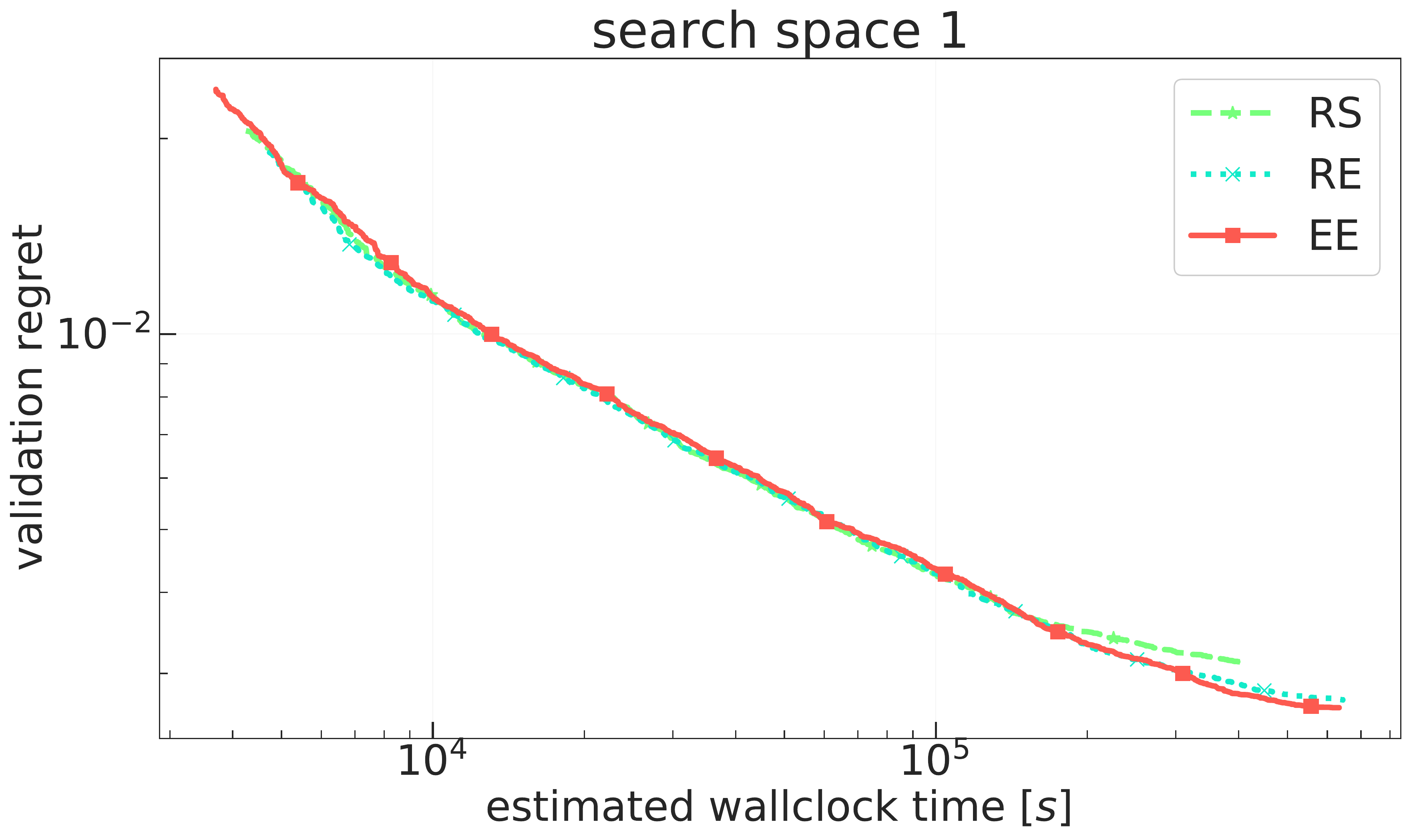}}
\hspace{\fill}
   \subfloat[\label{pyramidprocess} ]{%
      \includegraphics[width=0.3\textwidth]{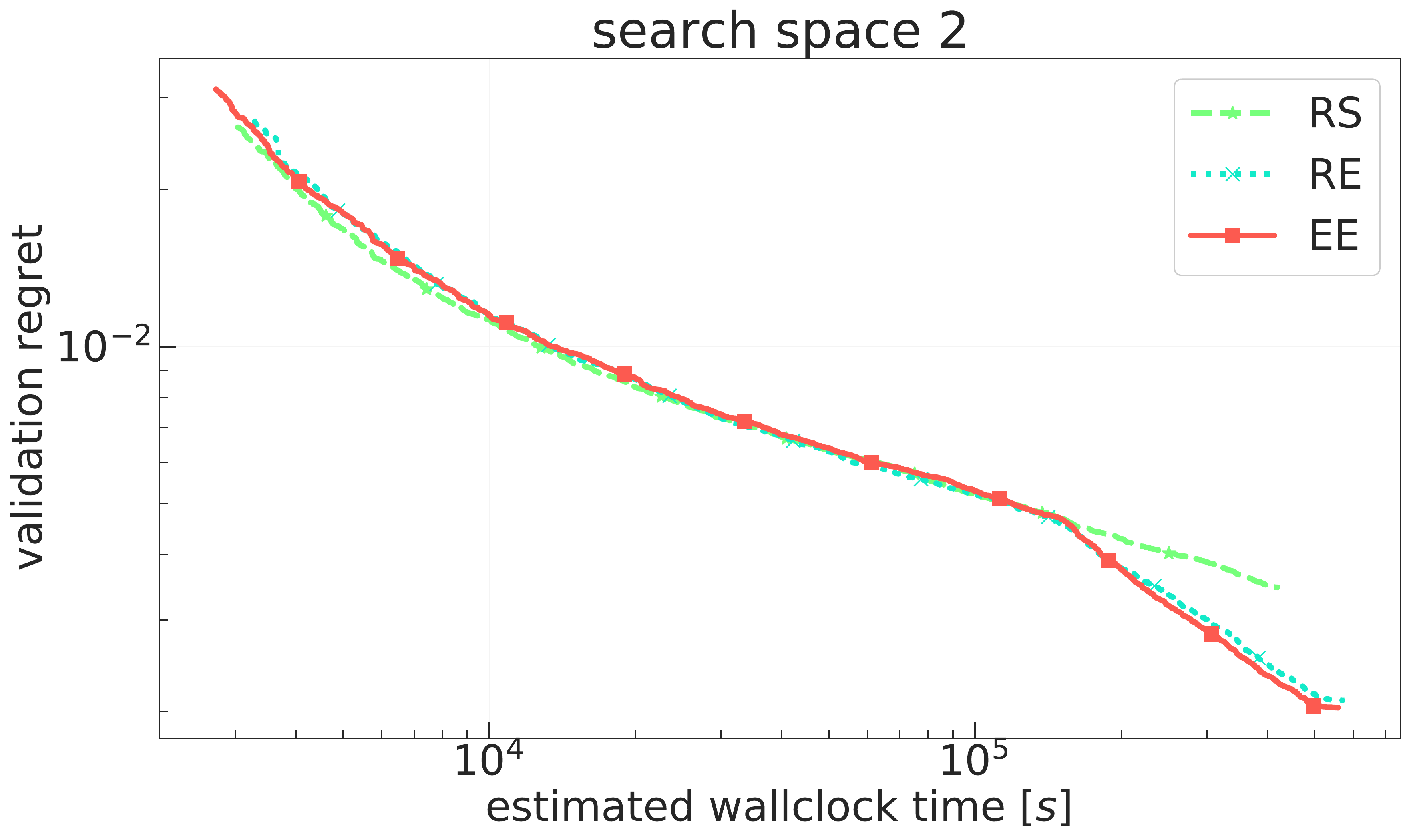}}
\hspace{\fill}
   \subfloat[\label{mt-simtask}]{%
      \includegraphics[width=0.3\textwidth]{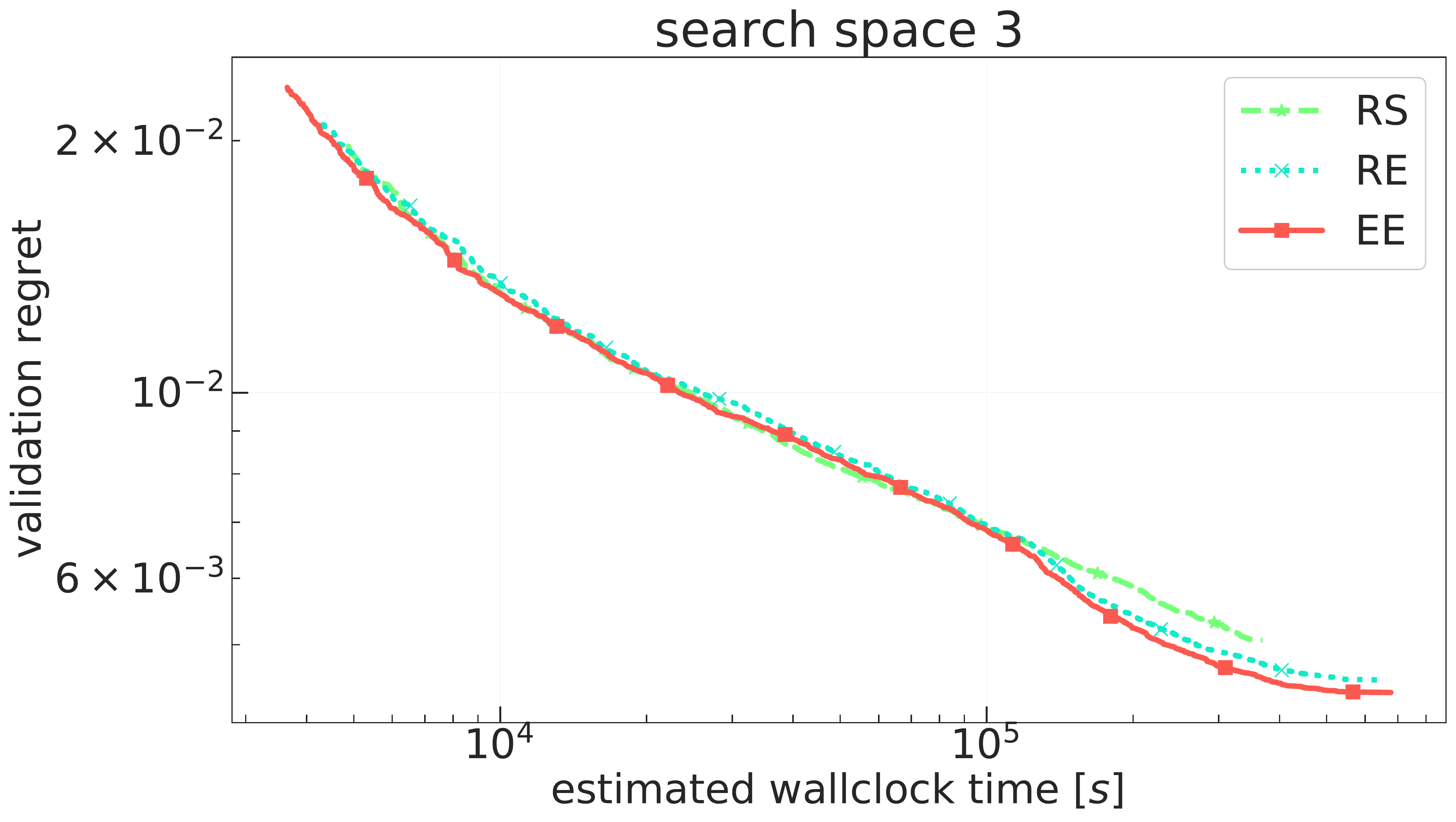}}
\caption{Comparison of accuracy between random search (RS), regularised evolution (RE) and Early Exit evolution (EE) algorithms on NAS-Bench-1Shot1.}
\label{fig:16}
\end{figure}

Fig.~\ref{fig:16} shows the mean performance on validation regret of architectures obtained by random search, regularised evolution and Early Exit evolution algorithms. For search space 1, our algorithm achieves validation regret close to the regularised evolution algorithm. For search space 2, our algorithm converges better than the regularised evolution algorithm. Our algorithm outperforms random search and regularised evolution algorithms for search space 3, the most significant (100$\times$ more architectures than space 1, and 10$\times$ more than space 2).

\subsubsection{NAS-Bench-201}

NAS-Bench-201 \cite{dong2020nas} is an extension of NAS-Bench-101, which extends different search spaces, and it has a wide range of datasets, including CIFAR-10, CIFAR-100, and ImageNet-16-120. It contains 15,625 architectures by five operations, and 6-dimensional vectors indicate the operation in the cell. All architecture evaluated performance by validation and test sets on CIFAR-10, CIFAR-100, and ImageNet-16-120. 

\begin{table}[t!]
\centering{%
\begin{tabular}{lcccccc}
\hline
\multirow{2}{*}{\textbf{Method}} & \multicolumn{2}{c}{\textbf{CIFAR-10}} & \multicolumn{2}{c}{\textbf{CIFAR-100}} & \multicolumn{2}{c}{\textbf{ImageNet16-120}} \\
                                 & \textbf{validation}  & \textbf{test}  & \textbf{validation}   & \textbf{test}  & \textbf{validation}     & \textbf{test}     \\ \hline
ResNet \cite{he2016deep}              & 90.83      & 93.97      & 70.42       & 70.86       & 44.53      & 43.63      \\ \hline
RSPS \cite{li2019random}                & 84.16±1.69 & 87.66±1.69 & 45.78±6.33  & 46.60±6.57  & 31.09±5.65 & 30.78±6.12 \\
Reinforce \cite{williams1992simple}          & 91.09±0.37 & 93.85±0.37 & 70.05±1.67  & 70.17±1.61  & 43.04±2.18 & 43.16±2.28 \\
ENAS \cite{pham2018efficient}                & 39.77±0.00 & 54.30±0.00 & 10.23±0.12  & 10.62±0.27  & 16.43±0.00 & 16.32±0.00 \\
DARTS \cite{liu2018darts}              & 39.77±0.00 & 54.30±0.00 & 38.57±0.00  & 38.97±0.00  & 18.87±0.00 & 18.41±0.00 \\
GDAS \cite{dong2019searching}                & 90.01±0.46 & 93.23±0.23 & 24.05±8.12  & 24.20±8.08  & 40.66±0.00 & 41.02±0.00 \\
SNAS \cite{xie2018snas}                & 90.10±1.04 & 92.77±0.83 & 69.69±2.39  & 69.34±1.98  & 42.84±1.79 & 43.16±2.64 \\
DSNAS \cite{hu2020dsnas}               & 89.66±0.29 & 93.08±0.13 & 30.87±16.40 & 31.01±16.38 & 40.61±0.09 & 41.07±0.09 \\
PC-DARTS \cite{xu2020pc}            & 89.96±0.15 & 93.41±0.30 & 67.12±0.39  & 67.48±0.89  & 40.83±0.08 & 41.31±0.22 \\ \hline
EA-Net (SO)         & 91.53±0.00 & 94.22±0.00 & 73.13±0.00  & 73.17±0.00  & 46.32±0.00 & 46.48±0.00 \\
EA-Net ($\beta$= 0)     & 88.97±2.48 & 91.54±2.69 & 66.84±5.08  & 67.00±4.90  & 39.93±5.54 & 39.27±6.21 \\
EEEA-Net ($\beta$= 0.3) & 87.07±1.59 & 89.76±1.87 & 64.04±3.21  & 64.31±3.21  & 35.42±3.81 & 34.98±4.13 \\
EEEA-Net ($\beta$= 0.4) & 89.91±0.77 & 92.68±0.69 & 68.70±1.50  & 68.65±1.51  & 41.71±1.58 & 41.25±1.61 \\
EEEA-Net ($\beta$=0.5) & 90.21±0.58 & 92.83±0.46 & 69.15±1.36  & 68.95±1.25  & 42.14±1.14 & 41.98±1.22 \\ \hline
Optimal             & 91.61      & 94.37      & 73.49       & 73.51       & 46.77      & 47.31      \\ \hline
\end{tabular}%
}
\caption{Comparison of a single objective and multi-objective evolution algorithm with the 8 \gls*{nas} methods provided by NAS-Bench-201 benchmark. Optimal shows the best architecture in the search space.}
\label{tab:9}
\end{table}

We compare our Early Exit Evolution Algorithm or EEEA-Net ($\beta$ = 0.3, 0.4 and 0.5) with the single objective evolution algorithms (SO) and multi-objective evolution algorithms ($\beta$ = 0). The hyper-parameters for this search process were defined as the generations of \gls*{ea} equal to 10 generations with 100 populations by retaining a probability of 0.5, a mutation probability of 0.1, and Early Exit is the maximum number of parameters equal to 0.3, 0.4 and 0.5 million. 

The results are shown in Table~\ref{tab:9}, our EEEA-Net ($\beta$ = 0.4 and 0.5) outperforms EEEA-Net ($\beta$ = 0 and 0.3). However, the EA-Net (SO) using accuracy as the optimisation objective performed better than all EEEA-Nets.

Furthermore, when we compared our EEEA-Net ($\beta$ = 0.5) with 8 NAS methods, including RSPS \cite{li2019random}, Reinforce \cite{williams1992simple}, ENAS \cite{pham2018efficient}, DARTS \cite{liu2018darts}, GDAS \cite{dong2019searching}, SNAS \cite{xie2018snas}, DSNAS, and PC-DARTS \cite{xu2020pc}, we found that EEEA-Net ($\beta$ = 0.5) has an accuracy was higher than all other NAS method except for the Reinforce method.


 
\end{document}